\documentclass[final,5p,times,twocolumn]{cls/elsarticle}

\usepackage{lineno}

\usepackage[ruled, vlined]{algorithm2e}
\usepackage{makecell}
\usepackage{graphicx,subcaption}
\usepackage{float}
\usepackage{amsmath,amsfonts,amssymb,amsthm,bm} % Math packages
\usepackage{enumitem}
\usepackage{diagbox} 	% table
\usepackage{booktabs}	% table
\usepackage{multirow}
\usepackage{mathtools}  % \overbrace

\usepackage[font=small,labelfont=bf]{caption}   % set caption size

\usepackage{tikz}

\usepackage[colorlinks=true]{hyperref}

\newcommand*{\figref}[2][]{%
  \hyperref[{#2}]{%
    Fig.~\ref*{#2}%
  }%
}

\newcommand*{\tabref}[2][]{%
  \hyperref[{#2}]{%
    Table.~\ref*{#2}%
  }%
}

\newcommand*{\eqnref}[2][]{%
  \hyperref[{#2}]{%
    (\ref*{#2})%
  }%
}

\newcommand*{\algref}[2][]{%
  \hyperref[{#2}]{%
    Alg.~\ref*{#2}%
  }%
}

\newcommand*{\sectref}[2][]{%
  \hyperref[{#2}]{%
    Section.~\ref*{#2}%
  }%
}

\AtBeginDocument{\hypersetup{allcolors=cyan}}

\journal{}

\begin{document}

\begin{frontmatter}



\title{Maps for Autonomous Driving: Full-process Survey and Frontiers}


%\author[a,b]{Wenzhong Shi $^\dagger$}
%\ead{lswzshi@polyu.edu.hk}

\author[a,b]{Pengxin Chen}
\ead{pengxin.chen@connect.polyu.hk}

\author[c]{Zhipeng Luo*}
\ead{lzp2680@mnnu.edu.cn}

\author[d]{Xiaoqi Jiang}
\ead{jiangxiaoqi1@mychery.com}

\author[a]{Zhangcai Yin}
\ead{yinzhangcai@whut.edu.cn}

\author[e,f]{Jonathan Li, \textit{Fellow IEEE}}
\ead{junli@uwaterloo.ca}

%\author[a,b]{Yue Yu}
%\ead{yue806.yu@connect.polyu.hk}
%
%
%\author[a,b]{Daping Yang}
%\ead{daping.yang@connect.polyu.hk}

\address[a]{School of Resource and Environment Engineering, Wuhan University of Technology, Wuhan, China}
\address[b]{Department of Smart Driving Product, IAS BU, Huawei Technologies, Shanghai, China}
\address[c]{Key Laboratory of Data Science and Intelligence Application of Fujian Province University and School of Computer Science, Minnan Normal University, Zhangzhou, China}
\address[d]{Global Technology Innovation Center, Chery Automobile Co., Ltd, Shanghai, China}
\address[e]{Department of Geography and Environmental Management, University of Waterloo, Waterloo, Canada}
\address[f]{Department of Systems Design Engineering, University of Waterloo, Waterloo, Canada}

\cortext[cor1]{Corresponding author}

\begin{abstract}

Maps have always been an essential component of autonomous driving. With the advancement of autonomous driving technology, both the representation and production process of maps have evolved substantially. The article categorizes the evolution of maps into three stages: High-Definition (HD) maps, Lightweight (Lite) maps, and Implicit maps. For each stage, we provide a comprehensive review of the map production workflow, with highlighting technical challenges involved and summarizing relevant solutions proposed by the academic community. Furthermore, we discuss cutting-edge research advances in map representations and explore how these innovations can be integrated into end-to-end autonomous driving frameworks.

\end{abstract}


\end{frontmatter}

%\section*{Cover Letter:}
%Over the past five years, the authors have witnessed the development of maps used in the field of autonomous driving (AD). They evolve from early days of pre-collection by professional mapping vehicles, through today's online vectorized map generation, to the future end-to-end road perception. Academia is flourishing, meanwhile the industry keeps selecting the most commercially feasible methods. In this survey paper, the authors divide the map development into three stages: HD Map, Lite Map and Implicit Map. Besides, the latest topics and trends are also summarized to provide insights into future research directions.

%\linenumbers

%%%%%%%%% BODY TEXT
%-------------------------------------------------------------------------
\section{Introduction}

Maps have long been an essential component of driving, providing critical spatial information for navigation and decision-making. In human-driven scenarios, Standard-Definition (SD) maps, commonly referred to as navigation maps, are generally sufficient. These maps typically include road-level structure and traffic rules, offering guidance for basic routing without requiring detailed lane-level information. Human drivers depend primarily on their own perception and reasoning capabilities to navigate with the aid of SD maps.

However, in Autonomous Driving (AD), the requirements for map precision and richness increase substantially. Autonomous vehicles rely heavily on HD maps which provide rich, centimeter-level lane information, including road geometry, lane boundaries, traffic signals, semantic landmarks, and even road surface features. These HD maps act as a prior knowledge base, enhancing localization accuracy, supporting perception systems, informing prediction models, and enabling rule-compliant planning.

Despite their critical role, HD maps present considerable limitations. They are expensive to produce, often require manual annotations and specialized survey vehicles, and suffer from slow update cycles and high maintenance costs, thereby hinder scalability. In response, Lite maps have emerged as a promising alternative. Lite maps retain essential lane-level information but leverage crowd-sourced data and automated generation techniques to significantly reduce production costs and support more frequent updates, offering a more scalable and practical solution for real-world AD deployment.

A more recent evolution is the development of implicit map representations, aligned with the trend toward end-to-end autonomous driving systems. Instead of storing explicit geometric and semantic elements, Implicit maps encode spatial and behavioral priors directly into neural network representations. These learned representations can be integrated into perception, prediction, and planning modules, enabling differentiable processing and backpropagation, which are essential in joint learning systems. The goal of this paradigm is to develop map representations that not only support driving tasks but also promote more human-like, context-aware decision-making in autonomous vehicles.

\begin{figure*}[thtb]
  \centering
  \includegraphics[width=.9\linewidth]{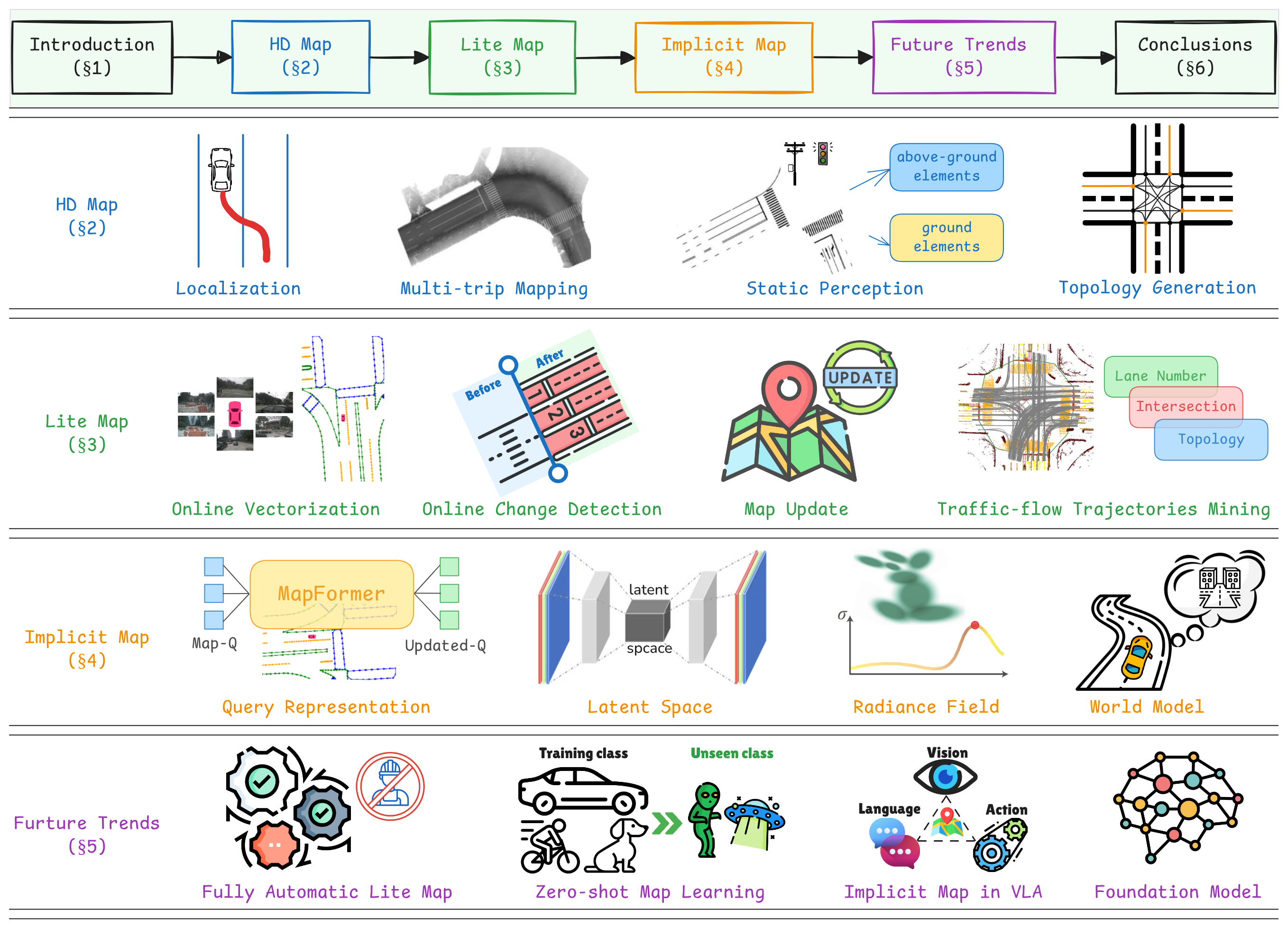}
   \caption{Survey at A Glance. The introduction section provides a technical overview of the modern autonomous driving and the map development. The structure of main body is organized into HD Map, Lite Map and Implicit Map according to the order of their appearance. The paper ends with summarizing future trends and conclusions.}
   \label{fig:paper-structure}
\end{figure*}

In this survey, we provide an comprehensive review of the maps used for autonomous driving. \figref{fig:paper-structure} provides an overview of our work. We begin by introducing the integration of maps in autonomous driving systems and the development of maps in industry. The main body is organized into four core sections: HD Map, Lite Map, Implicit Map, and a discussion on future trends. This arrangement is according to the chronological order of map development. In each development stage, we address the challenges encountered and review related works.

\begin{enumerate}[label=\arabic*)]
\item
HD Map Section: Focuses on localization, multi-trip mapping, static perception, and topology generation.
\item
Lite Map Section: Highlights online vectorization, change detection, map updates, and traffic-flow trajectory mining.
\item
Implicit Map Section: Explores different implicit representations and world models.
\item
Future Trends Section: Discusses emerging directions in the field, including fully automatic Lite maps, zero-shot map learning, vision-language-action models and foundation models.
\end{enumerate}

Through this structured organization, we aim to provide a thorough and insightful overview of modern mapping technologies and their evolving role in autonomous driving.

\subsection{Integration of Maps in AD Systems}

\begin{figure}[thtb]
  \centering
  \includegraphics[width=1\linewidth]{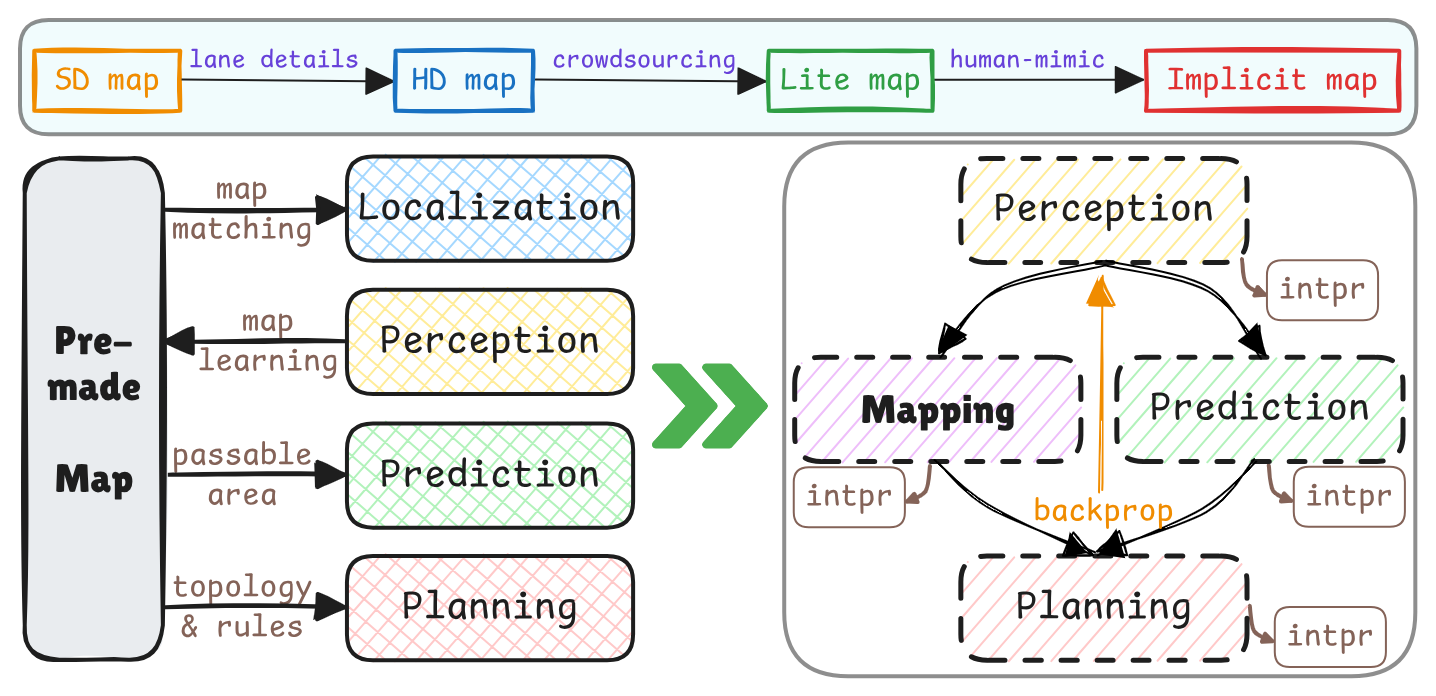}
   \caption{An overview of the evolution and integration of maps in autonomous driving systems. The top row illustrates the progression from \textbf{SD map}, sufficient for human drivers, to \textbf{HD map} with detailed lane information, followed by \textbf{Lite map} enabled through crowdsourcing, and finally to \textbf{Implicit map} designed for human-like, end-to-end autonomous driving. The bottom half contrasts the \textbf{\textit{traditional modular architecture}} (left) where pre-made maps independently support localization, perception, prediction, and planning with a \textbf{\textit{modern learning-based paradigm}} (right). On the right side, a shared Mapping module is jointly trained with other components via interpretable representations and backpropagation (marked as ``intpr'' and ``backprop'', respectively), allowing tighter integration and greater adaptability.}
   \label{fig:intro-archi}
\end{figure}

Maps have consistently played a vital role in AD architectures. \figref{fig:intro-archi} illustrates how maps are integrated into convention and modern AD systems, respectively.

In conventional modular AD architectures, each processing step, including the localization, perception, prediction, and planning, interacts with a prebuilt map in a predefined manner. For instance, map matching can help realize accurate localization in GNSS-denied environments. The prediction module relies on map information, along with the historical trajectories of vehicles or pedestrians, to forecast their future movements. Similarly, the planning module depends on the map’s topology and traffic rules to generate appropriate control strategies for vehicles.

In contrast, modern learning-based AD systems employ an end-to-end network architecture where multiple modules are optimized jointly. Within this framework, map representations are often integrated into a differentiable form, supporting gradient-based optimization and continuous feature learning across perception, prediction, and planning. This approach aims to improve coordination among functional components and ultimately achieve more human-like driving behavior. That said, such end-to-end systems have not yet reached the level of maturity required for large-scale commercial deployment.

Throughout the evolution of autonomous driving, maps themselves have also undergone significant transformation: moving from high-precision LiDAR-based surveying toward crowd-sourced data collection, from prebuilt offline maps to online map detection, and from explicit geometric representations to implicit neural embeddings. Regardless of these shifts, maps remain an indispensable element in AD systems—never absent, always evolving.

\subsection{Map Development in Industry}

\begin{figure*}[thtb]
  \centering
  \includegraphics[width=.8\linewidth]{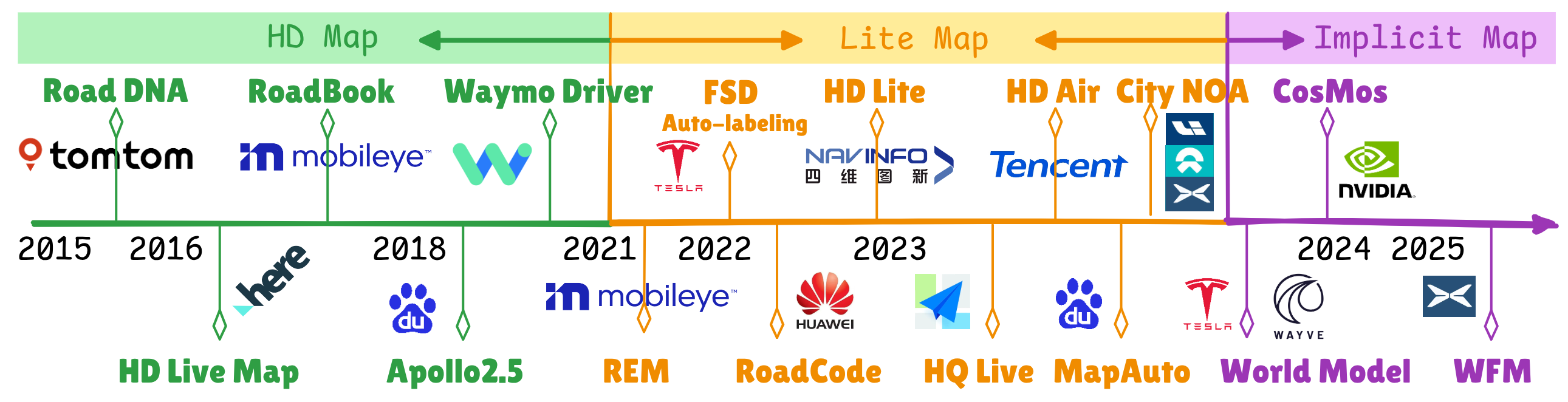}
   \caption{Key companies, technologies, and milestones from 2015 to 2025, illustrating the industry's transition from high-definition maps to lightweight maps, and now to implicit, AI-driven models.}
   \label{fig:intro-roadmap}
\end{figure*}

Based on the authors' industrial experiences for the last decade, we divide the development of maps for AD into three stages as shown in \figref{fig:intro-roadmap}.

\begin{enumerate}[label=\arabic*)]
\item
\textbf{HD Map Stage (2015–2021):} TomTom pioneered the first commercial HD map in 2015  \citep{tomtom2015roaddna}, establishing a critical foundation for map-dependent autonomous driving systems. During this phase, industry efforts concentrated on creating highly detailed and semantically rich maps that included precise lane geometries, traffic signs, and road attributes. Major contributors included TomTom with Road DNA, HERE with HD Live Map, and Mobileye with RoadBook. Significant developments also came from Baidu's Apollo 2.5 and Waymo Driver. These systems relied extensively on pre-built HD maps to support high-accuracy localization and contextual decision-making. Because of the expensive production and maintenance pipeline, most HD map products only cover highways for cost considerations.

\item
\textbf{Lite Map Stage (2021–Future):} Beginning around 2021, the industry transitioned toward ``Lite map", namely the lightweight maps, which aimed to reduce the dependency on dense and static HD maps by introducing more scalable and adaptive solutions. Tesla pioneered this trend with its FSD auto-labeling solution \citep{tesla_ai_day_2022}, emphasizing real-time perception and automated annotation. This stage also saw the emergence of Mobileye REM \citep{mobileye_rem_maps_2021}, Huawei’s RoadCode, NavInfo’s HD Lite, Tencent’s HD Air, and Baidu’s MapAuto, all focusing on lightweight, more updateable maps. These maps balanced efficiency and functionality, enabling wider geographic coverage with less infrastructure overhead. It is worthwhile to note that Lite maps remain the mainstream of industry at present and is far from being replaced by Implicit maps. Because the maintenance costs have been greatly reduced, Lite maps can cover urban roads.

\item
\textbf{Implicit Map Stage (2023–Future):} Since approximately 2023, a new trend has emerged toward Implicit maps, in which environmental knowledge is encoded implicitly within neural network parameters rather than through traditional explicit map formats. Mapping becomes an internalized process through AI large-scale learning. Rather than relying on explicit maps, vehicles understand their environment through learned models. Tesla and Wayve’s World Models \citep{tesla_cvpr2023_world_model, wayve_gaia1_2023}, NVIDIA’s CosMos \citep{agarwal2025cosmos_world} and Xpeng’s WFM \citep{xpeng_cvpr2025_world_model} represent this stage. These approaches aim to achieve highly generalizable and scalable autonomy by integrating map knowledge into end-to-end AD architectures.

\end{enumerate}

\subsection{Comparison to Related Surveys}

This survey distinguishes itself from prior works in two key aspects.

Firstly, while existing surveys \citep{elghazaly2023high, li2024development, asrat2024comprehensive, abdelkader2023hd, liu2020high} offer limited emphasis on industrial perspectives, this review incorporates insights grounded in real-world applications. As autonomous driving technology rapidly evolves toward production vehicles, distinct challenges emerge at different developmental stages. Accordingly, we adopt a dynamic, industry-informed viewpoint and categorize map evolution into three milestone phases: 1) HD Map; 2) Lite Map; and 3) Implicit Map. This staging allows us to systematically examine the unique technical challenges and innovations characteristic of each period.

Secondly, with regard to scope, previous surveys often concentrate on isolated topics, such as localization \& mapping \citep{wijaya2024high, charroud2024localization}, object detection \citep{luo2023road}, lane topology reasoning \citep{yao2025concise}, crowdsourcing update \citep{guo2024review, boubakri2022high} and end-to-end generation \citep{kwag2024review}, or focus exclusively on specific HD map production pipelines \citep{chen2023milestones, lyu2024online, bao2022high, tang2023high}. In contrast, our work provides a comprehensive full-process survey that spans the entire map production workflow, covering all essential modules from data collection to representation and application.

\subsection{Contributions}
The main contributions of our survey are as follows:
\begin{enumerate}[label=\arabic*)]
\item
We present a comprehensive review of map representations for autonomous driving. To the best of our knowledge, this is the first survey that systematically covers the entire developmental pipeline across three key stages: HD maps, Lite maps, and Implicit maps.

\item
We identify and analyze critical challenges faced at each stage from an industrial perspective. To support this analysis, we examine more than 200 relevant publications, offering insights grounded in both technical and practical contexts.

\item
We summarize emerging trends in Implicit maps and their role in end-to-end autonomous driving systems. Although most of these approaches have not yet been widely deployed commercially, we emphasize their potential to shape future autonomous driving models and encourage the community to actively engage with these developments. It is our hope that this survey will serve as a valuable reference and provide a holistic outlook for researchers and practitioners in both academia and industry.
\end{enumerate}

\section{HD Map: The First Victory on Highways}
\label{sect:hd}

	HD maps are characterized by lane-level precision and centimeter-level accuracy, making them a foundational infrastructure for autonomous driving systems. However, HD maps also face significant challenges, including high production costs and low update frequency. \textit{As a result, the early AD solutions which depends on HD maps could only cover highways.} This section elaborates on the HD map production pipeline and its key technical components.

\subsection{Production Pipeline}

\begin{figure}[thtb]
  \centering
  \includegraphics[width=.9\linewidth]{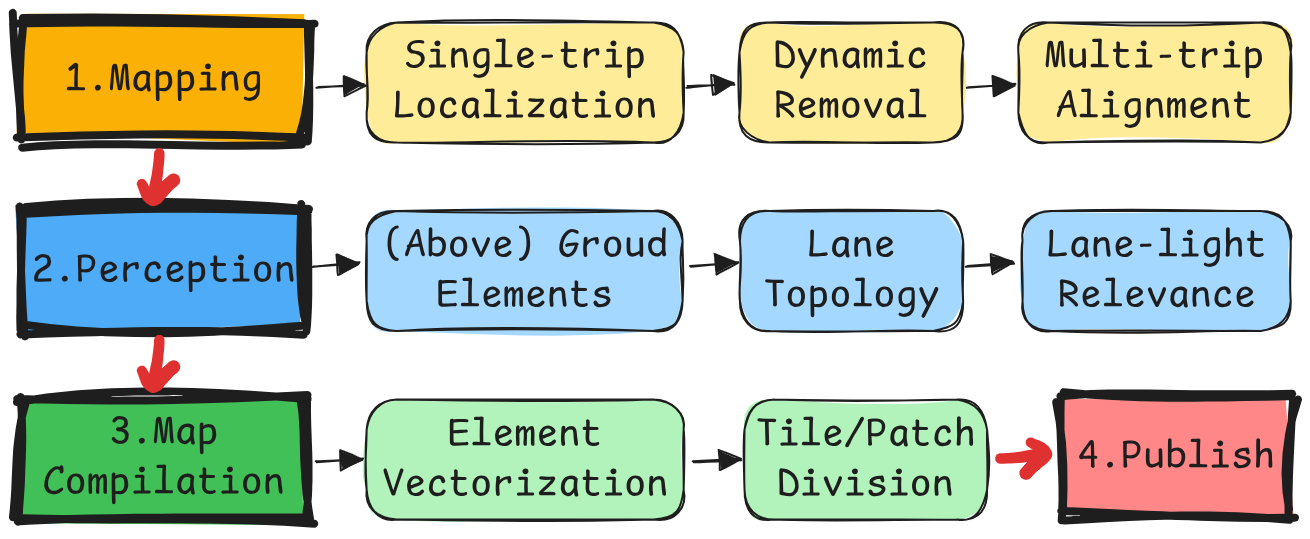}
   \caption{Typical pipeline of HD map production}
   \label{fig:hd-pipeline}
\end{figure}

We divide the production pipeline of HD map into three stages: Mapping, Perception and Map Compilation, as illustrated in \figref{fig:hd-pipeline}.

\begin{enumerate}[label=\arabic*)]
\item
\textbf{Mapping:} This stage aims to construct a globally consistent point cloud map by integrating data from multiple survey trips. For a survey vehicle, the location of each timestamp can be computed by integrated navigation or SLAM methods. After that, a point cloud map can be acquired by stitching laser scans. During this process, dynamic objects, such as pedestrains, cyclists and cars, are filtered out to retain only static structures. However, due to nonlinear errors and drift in localization, point cloud maps surveyed from different trips at the same place usually cannot be aligned well. To solve this problem, optimization-based algorithms are employed to align maps from different trips. Finally, the mapping process outputs a clean point cloud map as well as street-view images with absolute locations.

\item
\textbf{Perception:} As the downstream process of mapping, this stage involves extracting map elements essential for autonomous driving from raw point cloud maps and images. Both entity elements (e.g., ground features and above-ground objects) and logical elements (e.g., lane topology and traffic light associations) are processed in parallel. A detailed taxonomy of these elements is provided in \sectref{sect:road-elements}.

\item
\textbf{Map Compilation:} The extracted elements require further processing to support downstream applications such as planning and control (PnC), simulation, and map rendering. Map platforms compile these elements into structured, query-efficient vector representations. For instance, road elements are vectorized and spatially partitioned to facilitate efficient storage and retrieval (see \figref{fig:admin-division}).

\end{enumerate}

\subsection{Map specifications}
\label{sect:map-spec}

\subsubsection{File management}
\label{sect:hd-file}

\begin{figure}[thtb]
  \centering
  \includegraphics[width=.9\linewidth]{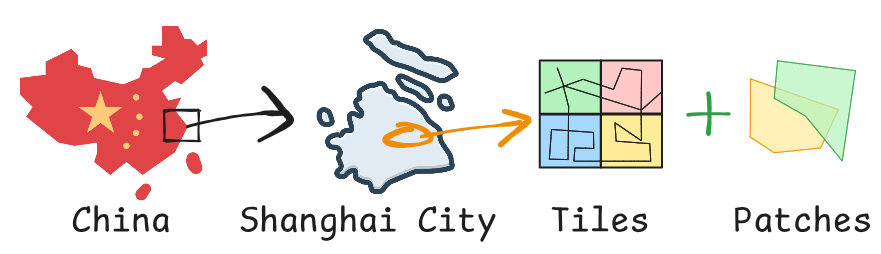}
   \caption{Map artifacts are managed by administrative divisions}
   \label{fig:admin-division}
\end{figure}

Before discussing the elements in HD maps, we would like to introduce file management. As illustrated in \figref{fig:admin-division}, map artifacts are usually stored and read by administrative divisions. Self-driving cars need to download the HD map portfolio belonging to their city. When a car starts navigating, both SD maps (i.e. navigation map) and HD maps are loaded into the memory. Since a HD map of the entire city is too large to be loaded to the memory all at once, the common practice is to only load the map data along the navigation route. Therefore, smaller scales of map division are required. 

\textbf{Tile} is a classic strategy to manage HD artifacts hierarchically. \cite{Tanner2004GeometryClipmaps} introduced a nested grid tile structure, known as geometry clip maps, for real-time terrain rendering. It efficiently supports high-resolution detail and smooth level-of-detail (LOD) transitions using a multilevel grid-based caching system. \cite{LokDuchaineauJoy2005TileHierarchies} applies hierarchical tile structures to planetary-scale terrain rendering. It introduces a “4–8” tile hierarchy, allowing efficient adaptive refinement of both geometry and texture using a pyramid-based tile organization. \cite{Zhang2016TilePyramidsVisualization} proposes a method for efficiently building tile pyramids from large-scale raster datasets. The generated multi-resolution tiles significantly accelerate visualization and interaction with massive geospatial data. \cite{WanHuangPeng2016NoSQLVectorTile} focuses on vector tile management using a hierarchical tile pyramid structure combined with NoSQL storage (HBase). It enables scalable and high-performance spatial indexing and retrieval for large vector datasets. \cite{Yan2023VectorTilePyramid} describes a distributed approach for constructing vector tile pyramids using a Hadoop-based platform. It uses a quadtree hierarchy to support scalable spatial connectivity analysis for agricultural land.

Tiles are hierarchical, compact, rectangular, friendly for indexing and retrieval, but they also have a disadvantage. That is, the tile representation mixes multi-level roads within a tile. More specifically, in complex interchanges, roads within a single connected region may exhibit vertical overlaps in the 2D projection. For instance, in downtown area, there may be an elevated bridge and an underground tunnel at the same place, so the road network will be interlaced vertically. These overlapping scenarios pose significant challenges for the downstream PnC module because we all known that it operates on planar road networks. 

To address this issue, it is necessary to further split a connected region into multiple patches according to the vertical stacking relationships of roads. The goal is to ensure that each resulting patch contains no overlapping structures in 2D, while satisfying the following constraints:
\begin{enumerate}[label=\arabic*)]
\item
Graph connectivity: The road network within each patch must remain topologically connected.
\item
Production specifications: Requirements such as maximum patch size, boundary completeness, and topological integrity must be maintained.
\end{enumerate}

\textbf{Patches} are irregular polygons. They are partitioned according to the ``no-interlacement'' rule. For example, if an elevated road is found above a main road, these two roads will be separated into two patches. The same rule applies to tunnels. According to the rule, the road network in a patch can be considered as a plane, thus facilitating downstream PnC. As the shapes of patches are irregular, they may overlap each other. A self-driving car needs to identify which patch it is in based on real-time locations and navigation routes.

\subsubsection{Elements}
\label{sect:road-elements}

\begin{figure}[thtb]
  \centering
  \includegraphics[width=.9\linewidth]{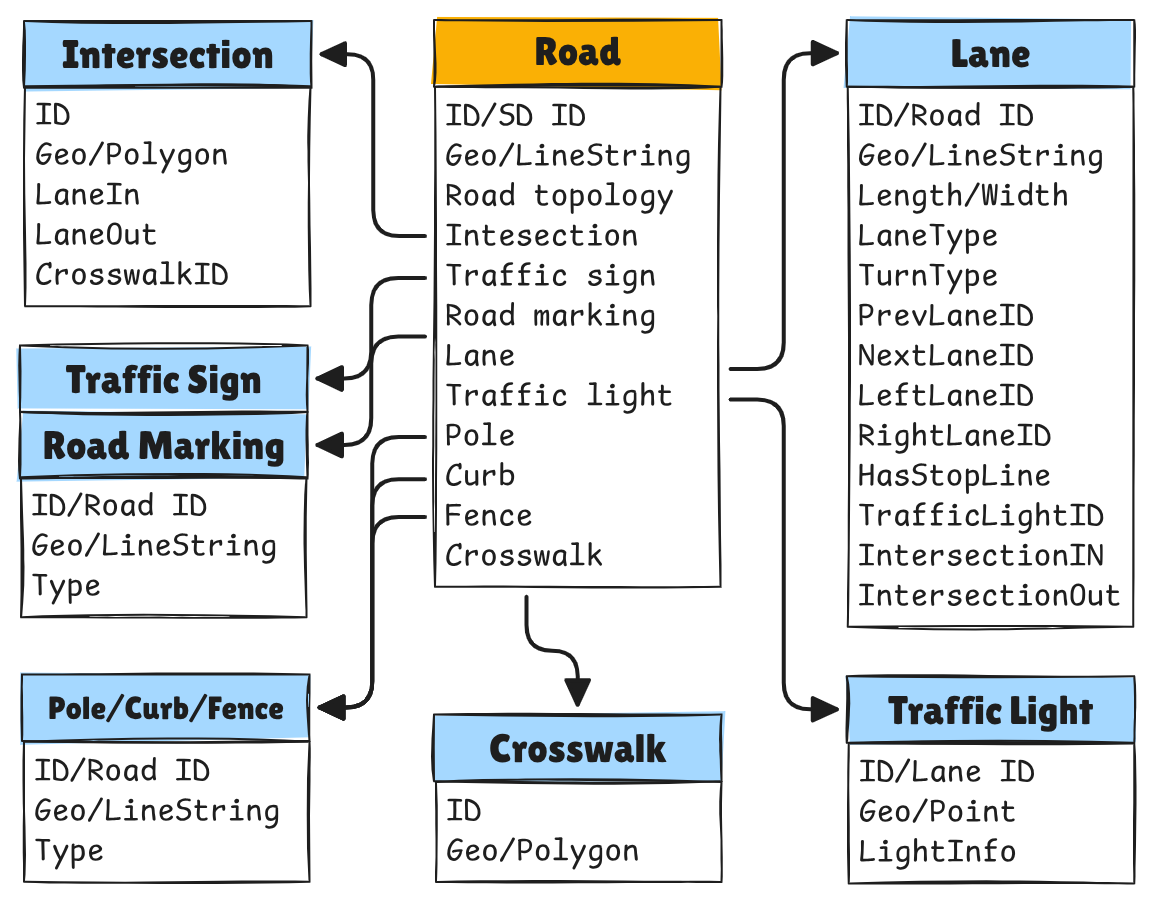}
   \caption{Typical elements in HD map}
   \label{fig:hd-elements}
\end{figure}

HD maps employ a structured, object-oriented schema to represent the complex relationships among diverse road elements. However, a unified standard has not yet been established across the industry, as different map providers often adopt varying specifications \citep{NDS2025, OpenDRIVE2023, poggenhans2018lanelet2, OpenCRG2014, guo2025refined}. To facilitate a clearer and systematic understanding, we distill the principal elements of an HD map into a simplified conceptual framework, illustrated in \figref{fig:hd-elements}. At the center lies the \textbf{Road}, which serves as the core entity and is linked to several sub-components.

\textbf{Road:} Each road is defined by a unique ID and its geometric representation (`Geo/LineString'). It also contains higher-level topology information and serves as a container for related objects such as traffic signs, road markings, traffic lights, lanes, poles, curbs, fences, and crosswalks.

\textbf{Lane:} A road may contain multiple lanes. Each lane is described by its geometry, length, width, and functional attributes such as `LaneType' and `TurnType'. The schema supports detailed lane-level connectivity through fields like `PrevLaneID', `NextLaneID', and lateral relationships (`LeftLaneID', `RightLaneID'). These refer to the IDs of neighboring lanes. The schema also includes indicators for stop lines, traffic light associations, and intersection connectivity.

\textbf{Intersection:} Intersections are modeled using polygonal geometries and contain references to incoming and outgoing lanes, as well as related crosswalks.

\textbf{Traffic Light:} Traffic lights are linked to lanes and defined by their spatial coordinates (`Geo/Point') and control information (`LightInfo'), enabling accurate modeling of signalized intersections.

\textbf{Traffic Sign} and \textbf{Road Marking}: These are associated with roads and described by type and geometry. They provide semantic cues for vehicle behavior and navigation.

\textbf{Pole/Curb/Fence:} These roadside objects are modeled as linear geometries and categorized by type. They contribute to the environmental context and are relevant for perception and localization tasks.

\textbf{Crosswalk:} Crosswalks are defined as independent polygonal objects but are also referenced by roads and intersections, aiding pedestrian-aware planning and safety modules.

\subsection{Localization}
\label{sect:hd-loc}
\begin{figure}[thtb]
  \centering
  \includegraphics[width=.6\linewidth]{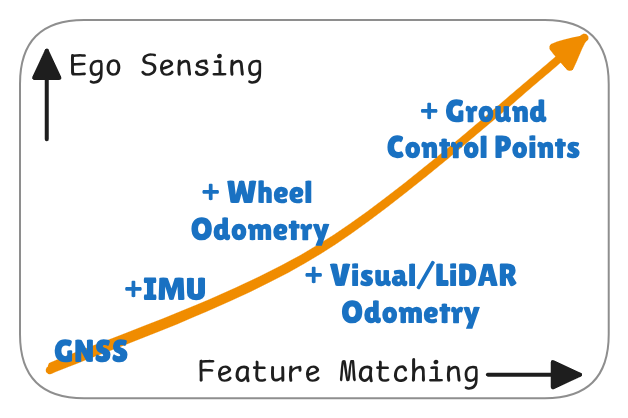}
   \caption{A schema of localization methods. The horizontal axis represents the increasing reliance on feature matching techniques, while the vertical axis reflects the advancement in ego-motion sensing capabilities.}
   \label{fig:hd-loc}
\end{figure}

Precise localization of survey vehicles is a fundamental component in the production of HD maps. These maps require centimeter-level precision to support Advanced Driver-Assistance Systems (ADAS) and autonomous driving. Survey vehicles equipped with high-precision GNSS, IMU, wheel odometer, LiDAR, and camera systems traverse road networks to collect rich environmental data. The effectiveness of this process hinges on the ability to precisely estimate the vehicle’s position and orientation. \figref{fig:hd-loc} illustrates the hierarchical enhancement of localization strategies employed in HD map surveying vehicles. 

\subsubsection{GNSS}
As the core component of positioning solutions, GNSS can provide accurate position and speed in open-sky areas. \cite{vandierendonck1992narrow} presents an early and influential analysis of the effect of correlator spacing on GPS receiver accuracy. It introduces the concept of “narrow correlator spacing”, which significantly improves pseudorange measurement accuracy, thereby enhancing both position and velocity estimation. The methodology laid a foundation for modern GPS receiver design. \cite{rife2004navigation} explores techniques for integrity monitoring in GNSS-based navigation, particularly for aviation. It evaluates the accuracy and robustness of position and velocity estimates derived from GPS, focusing on safety-critical applications. 

As the localization for HD map production does not require real-time processing, PPK (Post-Processed Kinematic) is preferred in many HD map manufactures such as NovAtel Inertial Explorer \citep{novatel_inertial_explorer}. Compared to RTK (Real-Time Kinematic), PPK does not require real-time data transmission. Therefore, PPK is more robust to data loss and allows full processing optimization. \cite{zhu2019multi_gnss_ppk} describes a robust framework for multi-GNSS PPK using double-difference carrier-phase processing. It improves ambiguity resolution by leveraging observations from multiple constellations, achieves centimeter-level precision, and analyzes ambiguity-fixing performance in post-processing mode. \cite{zhu2020improving_ppk} proposes strategies to improve integer ambiguity resolution in PPK by using quality control and optimized hypothesis testing. It enhances the ambiguity-fixing rate and robustness of the post-processing kinematic solution. 

\begin{figure}[thtb]
  \centering
  \includegraphics[width=1\linewidth]{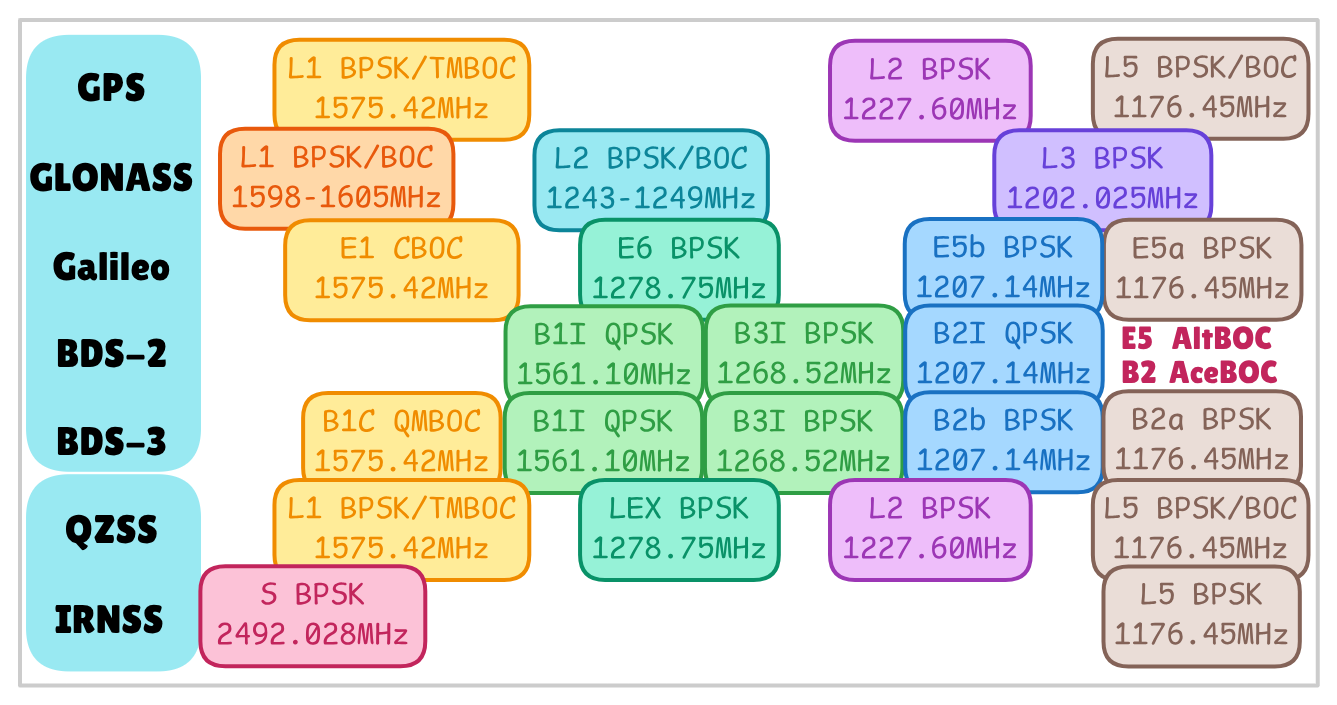}
   \caption{GNSS Signal Frequencies and Modulation Schemes. While all GNSS systems operate in similar frequency bands, each employs distinct modulation schemes to balance compatibility, precision, and robustness.}
   \label{fig:hd-gnss}
\end{figure}

Modern GNSS systems go beyond GPS and now include multiple constellations. Global GNSS systems with worldwide coverage include GPS (USA), GLONASS (Russia), Galileo (EU) and BeiDou (China). Regional systems include QZSS (Japan) and IRNSS/NavIC (India). \figref{fig:hd-gnss} provides an overview of these systems. 

\subsubsection{IMU}
IMU propagation can compensate for GNSS signal outages, enabling continuous positioning and smoothing the trajectory during GNSS-denied periods. \cite{Zhao2016_GNSSIMU} proposes a practical integration of a low-cost MEMS IMU with single-frequency GPS. It compares loose versus tight coupling strategies in dense urban scenarios. Using an Extended Kalman Filter (EKF), the authors analyze velocity/position error resilience during GNSS interruptions and highlighted drift behavior typical of MEMS-grade IMUs. \cite{Wen2020_FGOvsEKF} inspires a paradigm shift from extended Kalman filters to Factor Graph Optimization (FGO) in GNSS/INS navigation. The authors demonstrate FGO's superior accuracy during GNSS outages in urban canyons and discuss the influence of optimization window sizes.

\subsubsection{Wheel Odometry}
For vehicle localization, wheel odometry provides high-frequency motion updates that enhance short-term accuracy and stability. It is especially useful during GNSS outages or in environments where sensor visibility is limited, such as tunnels or urban canyons. \cite{zhang2021gnss} proposes a new navigation algorithm for all-wheel steering robots that combines GNSS, INS, odometry, and wheel angle data. It reduces positioning errors by using steering angles and speed to correct drift, even when GNSS is unavailable. Field tests show the method works well in both standard and all-wheel steering modes. \cite{mu2021data} addresses key issues in low-cost GNSS/INS/odometer fusion, such as odometer scale drift, sensor misalignment, lever arm effects, and wheel slip. It proposes solutions including scale and misalignment estimation, lever arm compensation, and fault detection using a two-stage Kalman filter. Results show improved positioning accuracy, especially in GNSS-denied environments. \cite{li2022high} presents a high-precision vehicle navigation system that tightly fuses GNSS PPP-RTK, MEMS IMU, wheel odometer, and vehicle motion constraints using a Kalman filter.

\subsubsection{Visual And LiDAR Odometry}
In extreme circumstances, such as a tunnel over 10KM long, the GNSS/IMU/Wheel Odometry fusion still suffers from the accumulation of drift errors. Feature matching provides inter-frame translation and attitude to solve this issue. In terms of sensor types, we classify the feature matching into 2 categories: \textbf{visual odometry (VO)} and \textbf{LiDAR odomtry (LO)}. VO captures texture features \citep{qin2018vins, campos2021orb}, while LO extracts geometry features \citep{zhang2017loam, chen2021backpack}. LO can acquire the real translation scale inherently, but VO needs to fuse the prior from IMU or binocular calibration to get it. In-P3VINS \citep{li2025p} proposes an invariant optimization-based system that fuses GNSS PPP, IMU, and Visual-SLAM for high-precision and consistent positioning. Built on an invariant factor graph framework, it models all raw measurements—including GNSS carrier phase with ambiguity states—for improved estimation. \cite{li2022tightly} proposes a tightly coupled GNSS PPP/INS/LiDAR integration system to meet the high-precision positioning demands of urban IoT applications like autonomous driving. It introduces a LiDAR sliding-window plane-feature tracking method to enhance both accuracy and efficiency. Field tests show the system achieves submeter-level accuracy in GNSS-challenging environments, significantly outperforming traditional GNSS/INS integration.

\subsubsection{Ground Control Points}
When neither texture nor geometry features are visible, both visual and LiDAR odometry will fail to localize itself. Ground control points (GCPs) provide fixed, precisely surveyed reference locations that anchor the HD map to real-world coordinates, ensuring accurate georeferencing. They can also help correct and reduce accumulated errors from GNSS, sensor drift, and mapping processes, thereby improving overall map reliability. They serve as ground truth for validating and calibrating HD map features, ensuring mapping quality meets stringent autonomous driving requirements. \cite{tsai2023alternative} aims to improve the efficiency of HD map production in Taiwan by using diverse surveying methods and multiple sources of ground control points (GCPs). The approach supports flexible and scalable map creation, allowing positioning within 0.5m and contributing to the evolution of HD map standards. \cite{chiang2022verification} outlines a verification and validation procedure for HD maps, emphasizing the role of GCPs in enhancing the relative and absolute accuracy of point clouds used in autonomous driving systems. The study highlights how GCPs contribute to meeting the stringent accuracy requirements of HD maps, ensuring reliable localization and navigation for autonomous vehicles.

\subsection{Multi-trip Mapping}
The localization methods introduced in \sectref{sect:hd-loc} are sufficient for the map construction of a single trip. Generally speaking, a single-trip map cannot cover all the lanes of a road due to sensors' range and environmental occlusion. Therefore, in practice, multiple trips are needed to make a complete map. However, different maps surveyed at the same place normally do not aligned well due to the different localization errors of them, thus leading to a scientific problem: multi-trip mapping. The problem is to aggregate the maps from multiple trips and ensure the mapping error is minimized.

Taking Huawei RoadCode as an example, a cloud mapping server is established to aggregate mass data captured by multiple vehicles. In 2021, the on-cloud mapping module is used to merge scalar maps uploaded by production cars \citep{qin2021light}. To save uploading bandwidth, the scalar map is designed in the form of a grid map whose resolution is $0.1 \times 0.1 \times 0.1 m$. Every grid’s information contains position, semantic labels, and counts of each semantic label. Semantic labels include ground, lane line, stop line, ground sign, and crosswalk. With the development of end-to-end online vectorized HD map generation, related work has shown decent speed and accuracy to be deployed onboard \citep{liu2023vectormapnet, liao2023maptr, yuan2024streammapnet, zhang2025mapexpert}. The vectorized form of a single-trip map seems to be a better choice to upload, taking into consideration of transmission bandwidth, computational overhead and surveying regulations. Therefore, in 2024, the on-cloud mapping module turns to directly consuming visual vectorized elements provided by the on-board cognition deployed on multiple cars \citep{chen2024mapcvv}. It then outputs global vectorized map which can be used for localization, planning and navigation. The new solution is more lightweight and efficient compared with its scalar-merging counterpart.

\subsubsection{Collaborative SLAM}
The multi-trip mapping task is quite similar to the collaborative SLAM defined in robotics literature. Taking self-driving cars as robots, the only difference between multi-trip mapping and collaborative SLAM might be that the former does not need real-time inter-robots communication. CoSLAM \citep{zou2012coslam} is the first visual-only SLAM solution in a dynamic environment with multiple cameras moving independently. This work introduces inter-camera pose estimation and inter-camera mapping to handle dynamic objects during localization and mapping, while maintaining uncertainty estimates for each map point to improve robustness. It also dynamically clusters cameras into groups based on view overlap, allowing real-time group splitting and merging, which enhances scalability and adaptability in dynamic environments. However, CoSLAM relies on the assumption that all cameras are synchronized and observe the same scene at initialization. Together with the requirement for a GPU, they render CoSLAM impractical to run online onboard multiple robots. CVI-SLAM \citep{karrer2018cvi} proposes a centralized collaborative SLAM framework which is more efficient. In the framework, multiple agents equipped with visual-inertial sensors share data with a central server, enabling efficient and robust multi-agent localization and mapping. It offloads computationally intensive tasks like global map optimization to the central server, reducing the onboard processing load while maintaining agent autonomy through onboard visual-inertial odometry. As a centralized framework, it inherently relies on a single server for data association and optimization, suffering from a communication bottleneck between the robots and the server. 

In contrast, decentralized solutions, such as DiSCo-SLAM \citep{huang2021disco} and Swarm-SLAM \citep{zhong2023dcl}, rely only on occasional communication between the robots are better suited for large-scale deployment. DiSCo-SLAM uses the Scan Context descriptor \citep{kim2018scan} in multi-robot LiDAR SLAM for data-efficient communication. The descriptor is lightweight and resilient to unknown initial conditions. Swarm-SLAM, a recent opensource multirobot SLAM system, presents a decentralized, scalable, and sparse collaborative SLAM framework specifically tailored for multi-robot (swarm) systems operating in GPS-denied environments. The system is modular and sensor-agnostic, supporting LiDAR, stereo, and RGB-D sensors, enhancing its flexibility across various robotic platforms. It also introduces a novel inter-robot loop closure prioritization strategy, which significantly reduces communication overhead and improves the speed of map convergence. Recently, there is a trend of using neural network to solve the collaborative SLAM problem. The CP-SLAM \citep{hu2023cp} is an example. It is the first dense collaborative neural SLAM framework based on a novel neural point based representation.

\subsubsection{Loop Closure Detection}
Loop closure detection and factor graph optimization are typically the back-end solvers used in multi-trip mapping. 

Some work focuses on the loop closure detection of visual images. \cite{angeli2008fast} proposes an online loop-closure detection approach by extending the bag-of-words image classification method with local shape and color features and incorporating Bayesian filtering to estimate loop-closure probability incrementally. The underlying idea is to compare a new image with all of the previous ones. If a loop closure is detected, it will be added to the back-end graph as an edge. Then, graph optimization is performed to share errors among edges according to customized weights. The size of the graph will gradually increase as the scene expands. If not controlled, the graph will be too large for real-time computation. To address this, a memory management mechanism, introduced by \cite{labbe2014online}, is used to limit the data processed by global loop closure detection and graph optimization. Thus, online constraints are independent of the size of the environment. As the prevalence of using 3D Gaussian Splat (3DGS) \citep{kerbl20233d} to represent 3D scenes, \cite{zhu2024loopsplat} presents LoopSplat, a dense RGB-D SLAM system that uses 3DGS submaps. The LoopSplat incorporates online loop closure through direct 3DGS registration and pose graph optimization for global consistency.

The LiDAR loop closure detection is also a significant task. \cite{xiang2021fastlcd} proposes FastLCD, a fast and compact loop-closure detection method that uses discriminative multimodal descriptors and a pre-trained machine learning classifier, followed by a double-deck verification strategy to improve accuracy. LCDNet \citep{cattaneo2022lcdnet} is a deep learning-based loop closure detection network for LiDAR point clouds, which combines a shared encoder, a place recognition head, and a novel relative pose head based on differentiable unbalanced optimal transport for end-to-end training. \cite{gupta2024effectively} proposes a loop closure detection approach for LiDAR SLAM that uses bird’s-eye view density images of locally generated maps, rather than individual scans, to create robust and compact descriptors suitable for long sequences and varying sensor patterns. It enables real-time loop closure detection with smaller feature complexity and database size.

Actually, there is an important trick in autonomous driving to solve the problem of loop closure detection. The trick is to match a car's continuous trajectory with the SD road networks that can be acquired from navigation maps. This is a well-defined task called map matching in navigation literature. It will be discussed in \sectref{hd-sd-match}.

\subsubsection{SD Map Matching}
\label{hd-sd-match}

Given SD road networks and a vehicle's moving trajectory, the goal of map matching is to project each point of the trajectory onto the road networks. The purpose is to recover the driving route of a car given its historical trajectory. Low localization accuracy and complex road topology are two difficulties to in this task.

The study of SD map matching can be traced back to \cite{bernstein1996introduction}. Now the solution paradigm has been basically determined. The paradigm uses Hidden Markov Model (HMM) to formulate the map matching problem and then applies Viterbi algorithm to solve the HMM. Most papers use this paradigm with some differences of formulating the distance from the given trajectory and road networks \citep{brakatsoulas2005map, bloit2008short, newson2009hidden, lou2009map, goh2012online}.

\subsection{Static Perception for HD Map Construction}

\begin{figure}[thtb]
  \centering
  \includegraphics[width=1\linewidth]{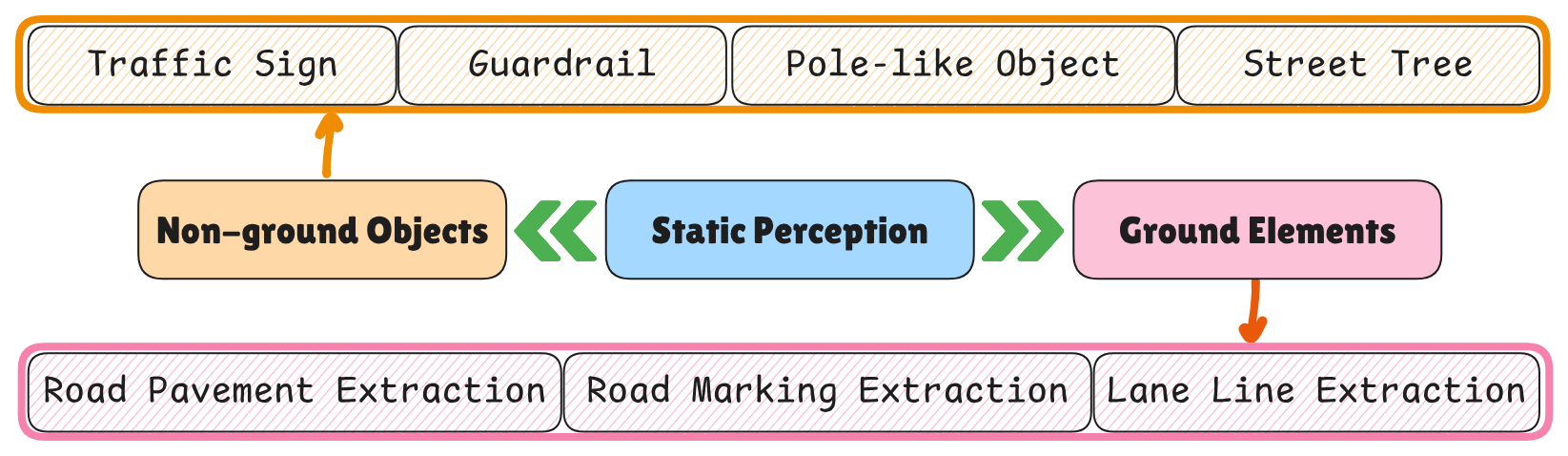}
   \caption{Ground elements detection and non-ground objects detection for static perception}
   \label{fig:hd-perception}
\end{figure}

Static perception is a fundamental stage in HD map generation, focusing on the extraction of both ground elements (encompassing pavement, markings, and lane lines) and non-ground objects, such as traffic signs, guardrails, pole-like structures, and street trees, as shown in \figref{fig:hd-perception}.

\subsubsection{Ground Elements Extraction}

\begin{figure}[thtb]
  \centering
  \includegraphics[width=1\linewidth]{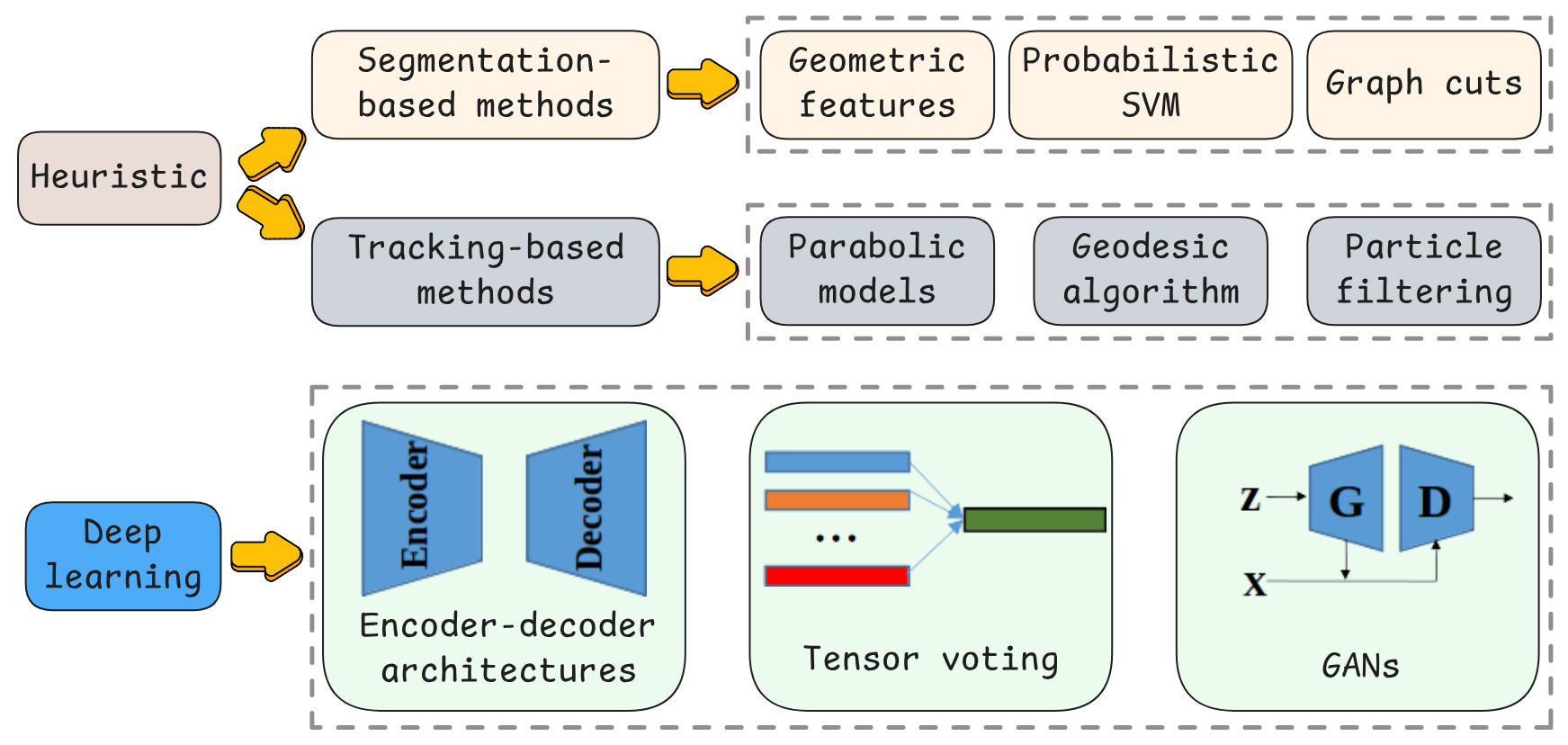}
   \caption{Two kinds of road pavement extraction methods}
   \label{fig:hd-pavement}
\end{figure}

\textbf{Road Pavement Extraction:} As shown in \figref{fig:hd-pavement}, early approaches primarily relied on heuristic ideas. Segmentation based techniques formulate the task as an image segmentation problem, frequently integrating geometric features \citep{hu2007road}, probabilistic Support Vector Machines (SVM), and graph cuts \citep{cheng2014urban}. Conversely, tracking-based methods connect seed points using parabolic models \citep{hu2004robust}, geodesic algorithms \citep{miao2014semi}, or particle filtering \citep{zhou2006road}. However, these methods often face limitations in complex urban scenes due to their dependence on empirical features and seed point initialization. Deep learning (DL) approaches have become the predominant solutions. Encoder-decoder architectures significantly enhance multi-scale feature learning. To address occlusion challenges, post-processing techniques including tensor voting \citep{gao2019road} and Generative Adversarial Networks (GANs) \citep{zhang2019aerial} are employed to reconnect fragmented road regions. Despite improved accuracy, effectively modeling topological relationships in heavily occluded areas, such as those under dense vegetation, remains challenging.

\begin{figure}[thtb]
  \centering
  \includegraphics[width=.65\linewidth]{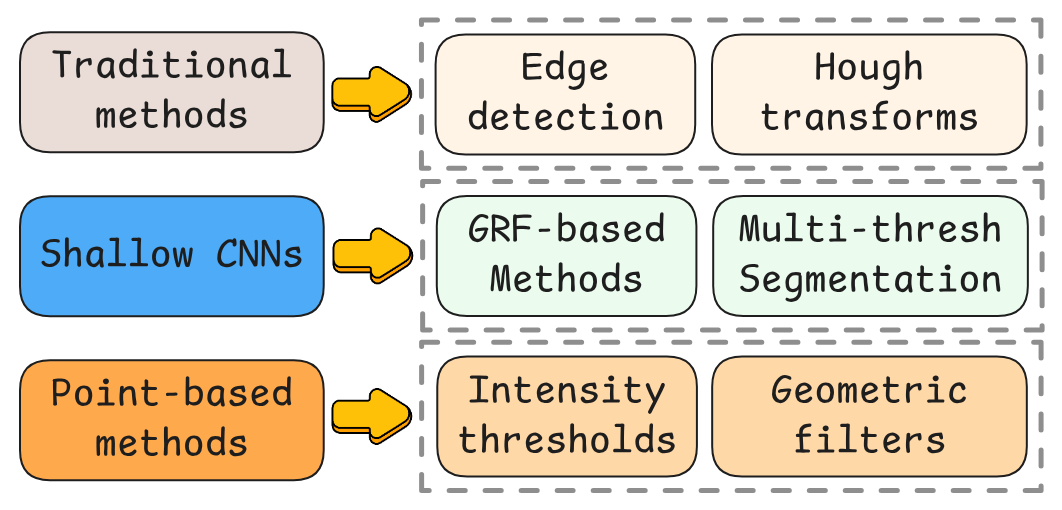}
   \caption{Three kinds of methods for road marking extraction}
   \label{fig:hd-roadmarking}
\end{figure}

\textbf{Road Marking Extraction:} As shown in \figref{fig:hd-roadmarking}, three primary strategies characterize this task. Traditional 2D vision-based methods exploit the inherent color contrast between markings and pavement, often employing edge detection or Hough transforms \citep{ying2016robust}. While shallow CNNs have been applied for marking classification \citep{bailo2017robust}, they exhibit significant performance degradation under varying illumination conditions and occlusions. Geo-Referenced Feature (GRF) Image-based Methods apply radiometric correction \citep{guan2014using} and multi-threshold segmentation to LiDAR intensity data projected onto 2D images. However, this projection inevitably causes information loss, frequently resulting in incomplete extractions. In contrast, point-based methods directly process 3D point clouds using intensity thresholds and geometric filters \citep{wen2019deep}, demonstrating superior robustness. Key limitations include difficulty in detecting worn markings with low reflectivity and susceptibility to environmental obstructions.

\textbf{Lane Line Extraction:} Traditional pipelines encompass several approaches. The geometric model typically involves edge detection followed by polynomial fitting \citep{mccall2006video}. The Hough Transform (HT) detects lines, which are subsequently refined using B-splines \citep{deng2013real}. The energy minimization technique uses conditional random fields to model multi-lane correlations \citep{hur2013multi}. Deep learning approaches currently dominate recent research. Segmentation-focused networks, including encoder-decoder CNNs \citep{kim2017end} and Spatial CNNs \citep{pan2018spatial}, excel at capturing long-range contextual features. Weakly supervised methods have also emerged, leveraging multi-sensor data to reduce the substantial costs associated with manual annotation \citep{bruls2018mark}. Future research may focus on integrating sequence modeling techniques, such as RNNs or Transformers, to further improve performance, particularly in challenging occlusion scenarios.

%\begin{figure}[thtb]
%  \centering
%  \includegraphics[width=.6\linewidth]{img/hd-laneline.png}
%   \caption{Traditional pipelines and DL-based methods for lane line extraction}
%   \label{fig:hd-laneline}
%\end{figure}

\subsubsection{Non-ground Objects Detection}

\begin{figure}[thtb]
  \centering
  \includegraphics[width=.95\linewidth]{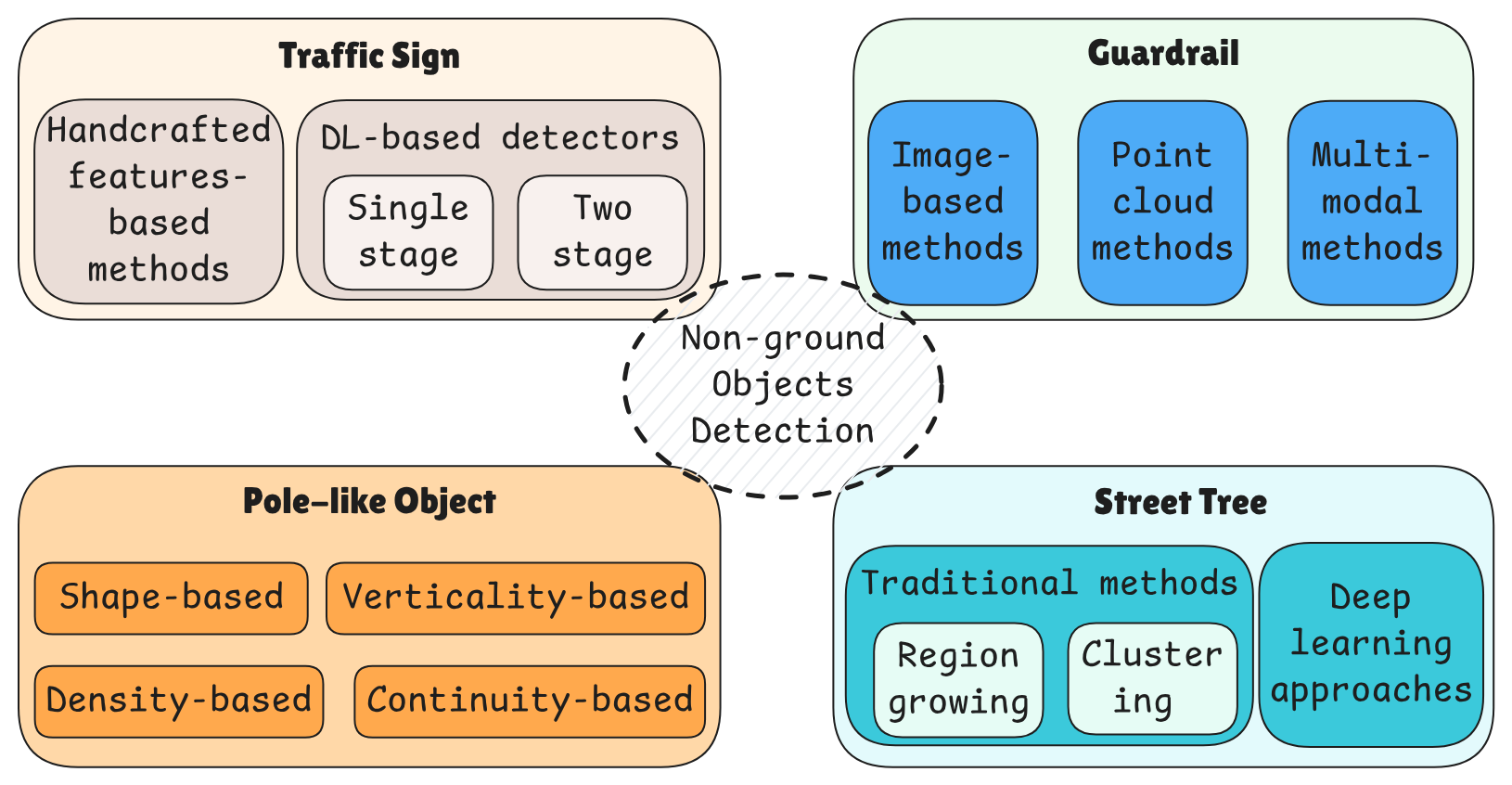}
   \caption{Four common categories in non-ground objects detection}
   \label{fig:hd-nonground}
\end{figure}

\figref{fig:hd-nonground} gives an overview of non-ground objects detection.

\textbf{Traffic Sign Detection:} Historically, image-based methods utilizing handcrafted features are prevalent. These include color/shape thresholding \citep{de1997road}, Maximally Stable Extremal Regions (MSER) \citep{salti2015traffic}, and HOG-SVM classifiers \citep{zaklouta2014real}. DL-based detectors have largely supplanted these approaches. Two-stage frameworks like Faster R-CNN \citep{ren2016faster} and Mask R-CNN \citep{he2017mask} achieve high accuracy but face efficiency challenges. Single-stage models, such as variants of YOLO \citep{redmon2016you}, effectively balance speed and performance \citep{avramovic2020neural}. Point cloud-based methods employ techniques like supervoxel segmentation \citep{yang2015hierarchical}, Hough forests \citep{wang2014object}, and Bag-of-Visual-Phrases combined with deep Boltzmann machines \citep{yu2016bag}. However, sparse point distribution and sensitivity to noise pose critical challenges for detection relying solely on point clouds. Consequently, fusion strategies incorporating image data represent a promising direction for enhancing overall robustness. 

\textbf{Guardrail Detection:} Image-based approaches, often utilizing stereo-vision systems with geometric constraints \citep{scharwachter2014visual}, exhibit limited reliability, particularly under nighttime conditions. Point cloud methods primarily rely on spatial clustering and feature-based filtering (e.g., corner features, height constraints) to detect both anti-fall and lane-separating guardrails \citep{gao2020rapid}, although their dependence on handcrafted features restricts generalization capability. Multi-modal fusion techniques, which combine image patches classified by CNNs with corresponding point clouds, demonstrate enhanced detection performance \citep{matsumoto2019extraction}.

\textbf{Pole-like Object Detection:} Methods for detecting pole-like objects can be categorized into four main types, as shown in \figref{fig:hd-nonground}. Shape-based approaches, such as RANSAC or least-squares fitting, identify cylindrical structures \citep{lam2010urban, huang2015pole}. Verticality-based methods exploit the characteristic vertical continuity of these objects in 3D space \citep{fukano2015detection, yan2017detection}. Density-based techniques project points onto 2D grids and apply clustering algorithms like DBSCAN \citep{yan2016automatic, zheng2016pole}. Continuity-based methods utilize voxel analysis to extract vertically connected components \citep{wu2013voxel}. While verticality/continuity methods generally offer higher accuracy, density-based approaches favor computational efficiency. A common limitation across all categories is their reliance on handcrafted features.

\textbf{Street Tree Detection:} Traditional methods focus on point cloud segmentation. Region growing algorithms utilize properties like curvature \citep{besl2002segmentation}, normals \citep{tovari2005segmentation}, or voxel merging \citep{vo2015octree} to isolate trees. Clustering algorithms, such as Mean Shift \citep{ferraz20123} and Normalized Cuts \citep{reitberger20093d}, are also employed to separate individual trees. More recently, deep learning approaches, particularly R-CNN frameworks augmented with attention modules, have shown promise in improving detection accuracy under occlusion conditions \citep{xie2019detecting}. A persistent challenge in segmentation arises from the spatial proximity of trees to streetlights and other urban structures \citep{rastiveis2020automated}.

\subsubsection{Summary}
The extraction of road surfaces is increasingly adopting DL-based segmentation techniques, yet significant challenges remain in preserving topological connectivity under heavy occlusion. Road marking extraction methods benefit from direct point cloud analysis but require enhanced robustness against degradation and environmental factors. Traffic sign detection leverages advanced 2D detectors effectively but necessitates further improvements in detecting small targets and developing robust multimodal fusion strategies. Detection of pole-like objects and street trees still relies heavily on traditional point cloud geometry analysis, warranting greater integration with advanced 3D deep learning paradigms. Guardrail detection demands the development of more adaptive feature learning frameworks. Two persistent overarching challenges remain across static perception tasks:
\begin{enumerate}[label=\arabic*)]
\item
Data Dependency: The labor-intensive nature of annotating 3D point clouds for training and validation remains a significant bottleneck. Promising solutions involve exploring synthetic data generation \citep{li2020deep} and transfer learning techniques to reduce annotation burdens.
\item
Unified Frameworks: Current pipelines typically process different map elements separately. There is an emerging need for end-to-end models capable of jointly extracting both ground and non-ground features \citep{mi2021hdmapgen} within a unified framework to improve consistency and efficiency.
\end{enumerate}

\subsection{Topology Generation}

\begin{figure}[thtb]
	\centering
	\begin{subfigure}[t]{.495\linewidth}
    \centering
    \includegraphics[width=1.\linewidth]{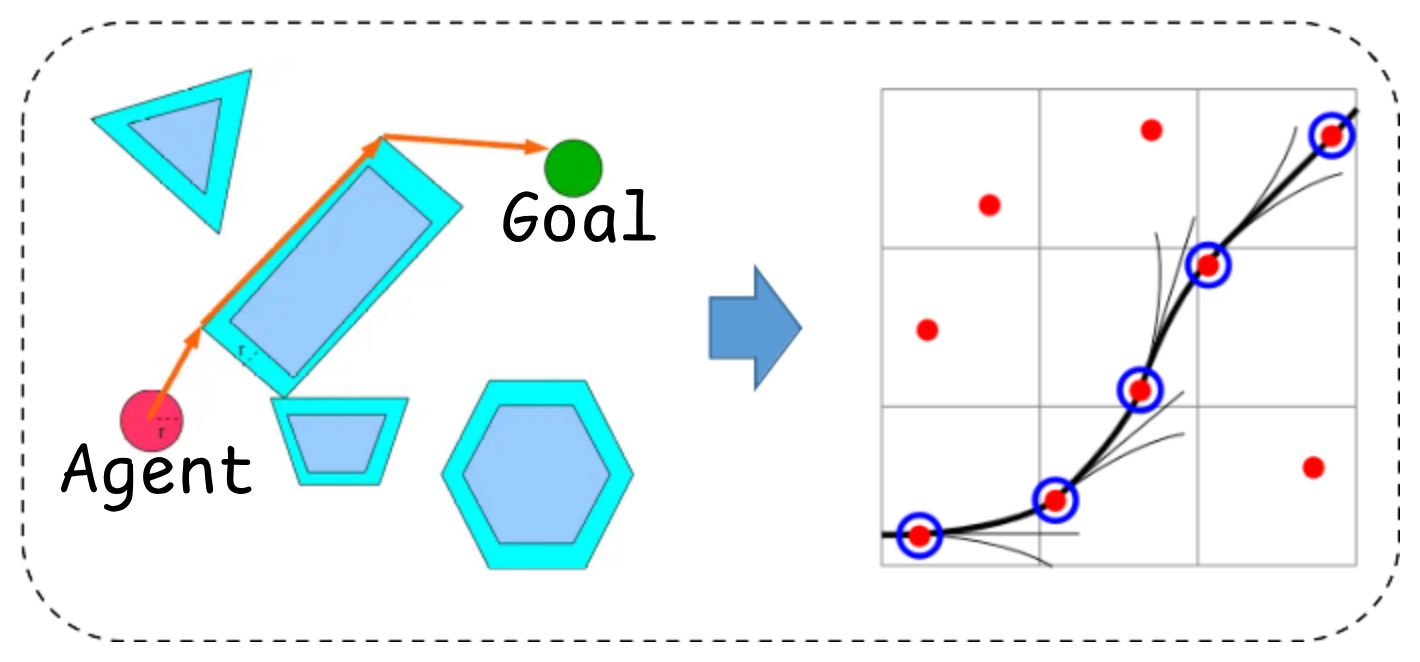}
    \caption{Voronoi-based methods}
    \label{}
  	\end{subfigure}	
	\begin{subfigure}[t]{.495\linewidth}
    \centering
    \includegraphics[width=1.\linewidth]{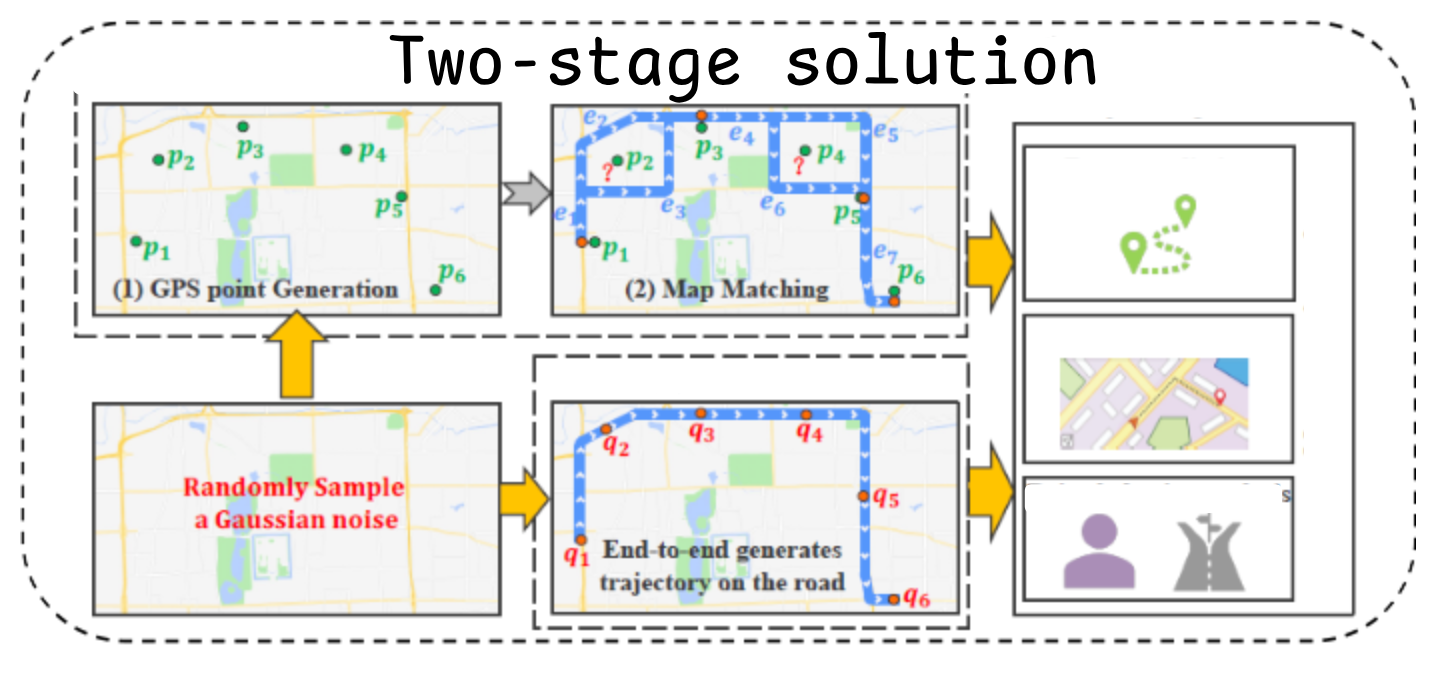}
    \caption{Deep Learning-based methods}
    \label{}
  	\end{subfigure}
	\begin{subfigure}[t]{.495\linewidth}
    \centering
    \includegraphics[width=1.\linewidth]{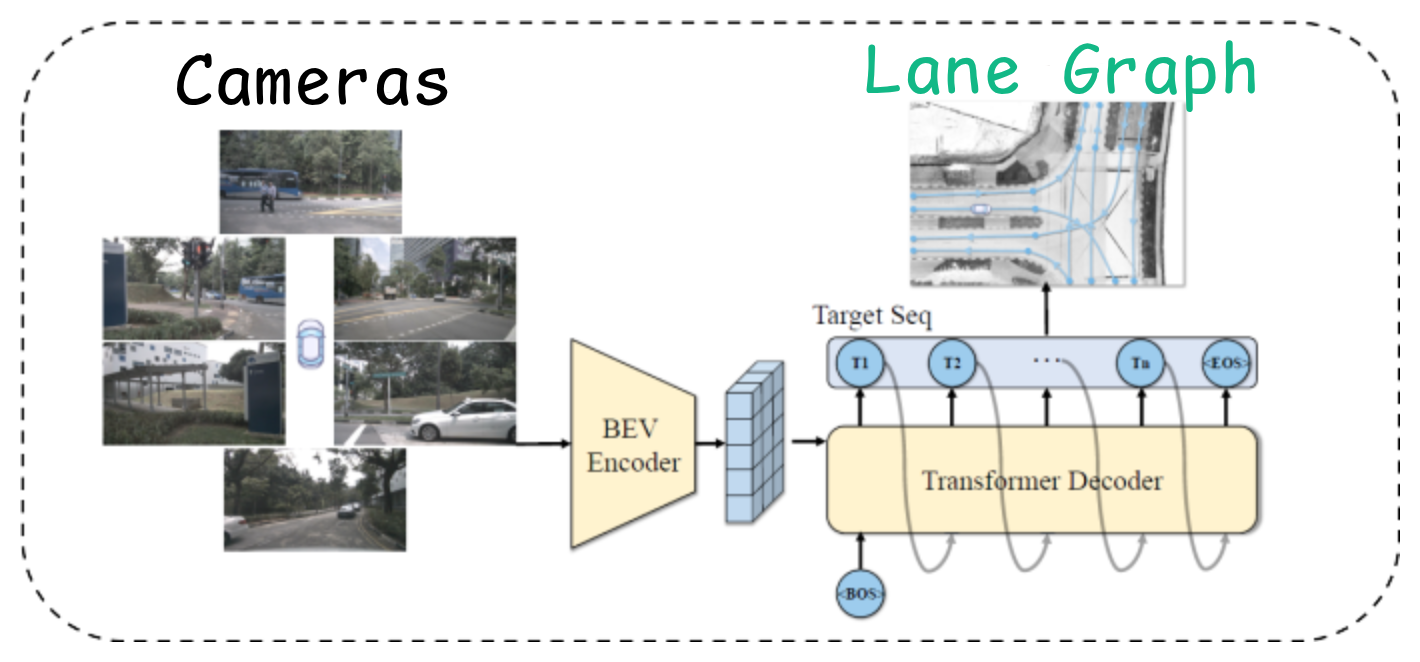}
    \caption{Hierarchical progressive methods}
    \label{}
  	\end{subfigure}	
	\begin{subfigure}[t]{.495\linewidth}
    \centering
    \includegraphics[width=1.\linewidth]{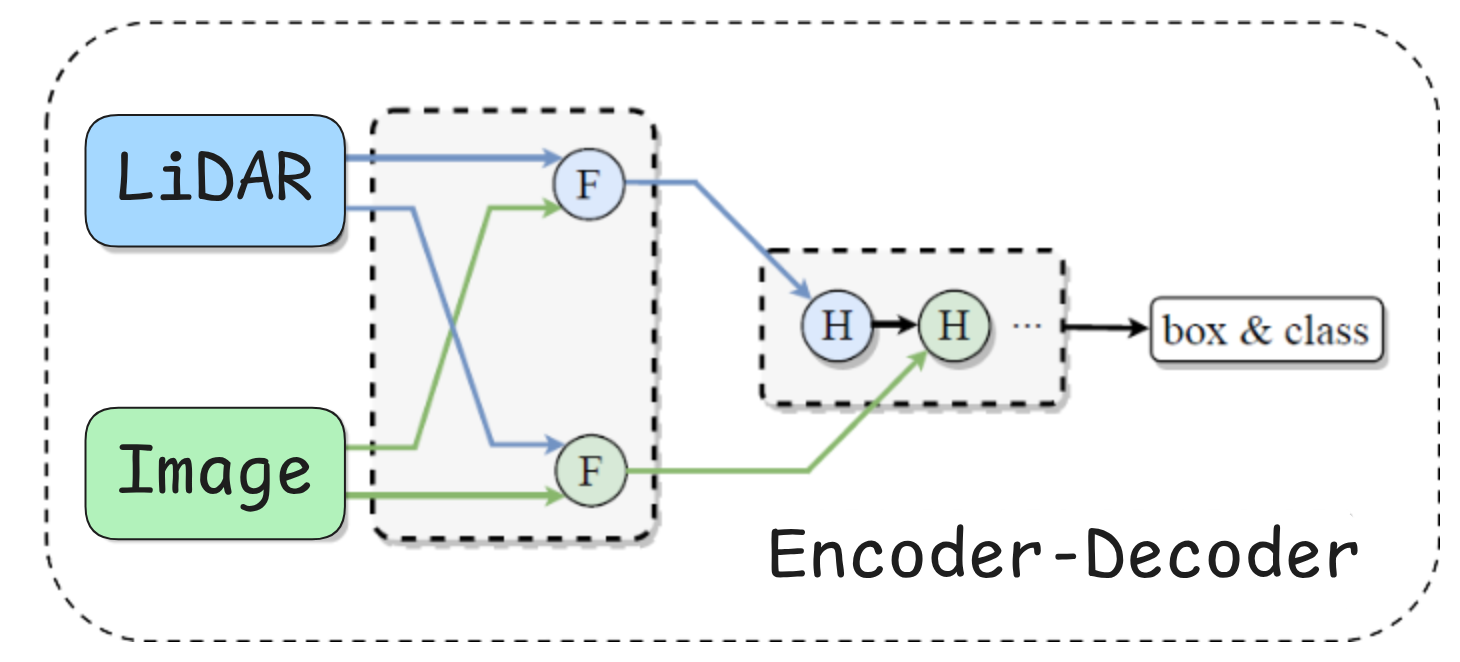}
    \caption{Multimodal Fusion methods}
    \label{}
  	\end{subfigure}
	\caption{Four categories of topologic generation methods}	
	\label{fig:hd-topo} 	
\end{figure}

As the "road genome" of autonomous driving systems, topological maps abstractly describe key nodes (e.g., intersections, lane points) and connection relationships (e.g., lane centerlines, traffic rules) in road networks. They serve as core elements for efficient path planning and navigation decision-making. Topological maps offer high storage efficiency, low computational burden, and easy dynamic updates, which makes them more suitable for real-time localization and environmental understanding in autonomous vehicles. With the advancement of autonomous driving technology, topology generation methods have evolved from traditional geometric algorithms to data-driven approaches. Based on different technical routes and core ideas, as shown in \figref{fig:hd-topo}, topology generation methods can be divided into four categories: graph search and geometric constraints methods, deep learning-based methods, hierarchical structures and progressive reasoning methods, and multimodal fusion and map prior-based methods. In this subsection, we analyze the core ideas, advantages, and limitations of each category, and provides prospects for future research directions.

\subsubsection{Graph Search and Geometric Constraints}
This category is grounded in graph theory and computational geometry. It achieves path planning by constructing a Configuration space (C-space) and a Generalized Voronoi Diagram (GVD). The core lies in transforming physical spatial obstacles into topological constraints and employing heuristic search algorithms to identify optimal paths. \cite{bhattacharya2012topological} proposed a product structure algorithm for high-dimensional configuration spaces, reducing complexity through topological decomposition, significantly enhancing the planning efficiency of multi-robot systems. \cite{sedighi2019implementing} introduces a Voronoi-guided hybrid A* algorithm, which integrates the topological separation characteristics of Voronoi diagrams with the Reed–Shepp kinematic model. \cite{liu2022path} offers theoretical rigor and provable safety. Its typical framework consists of three levels: environmental topology modeling, path space decomposition, and conflict avoidance, making it suitable for multi-agent collaborative navigation in structured environments. \cite{ho2022voronoi} employs direction-constrained search and real-time replanning mechanisms to control the collision rate in dynamic scenarios, addressing the limitations of traditional methods in responding to environmental dynamics.

Thought these methods have achieved satisfactory results in some scenes, they still face several challenges. Firstly, the accuracy of dynamic obstacle prediction depends on the completeness of environmental modeling. Secondly, the computational complexity of high-dimensional configuration spaces grows exponentially with the number of agents. Thirdly, generated paths require post-processing to meet vehicle dynamic constraints.

\subsubsection{Generative Models and Deep Learning}
Generative models learn the distribution patterns of topological structures in a data-driven manner, with their core advantage being the reduction of manual design costs and the generation of road networks that conform to real statistical characteristics. The typical workflow of these methods consists of three stages: data preprocessing, topological feature learning, and controllable generation, making them suitable for large-scale road network construction in autonomous driving simulation scenarios. Early approaches were represented by GANs, such as \cite{owaki2020roadnetgan}, while diffusion models have recently become mainstream, improving generation quality through latent space topological constraints. For example, structure-aware diffusion models \citep{wei2024diff} incorporate persistent homology theory to address the issue of fragmented generated trajectories, improving topological accuracy by 32\%. Recently, language-guided traffic simulation models \citep{zhong2023language} fuse Large Language Models (LLMs) with diffusion models, enabling road network generation controlled by natural language instructions.

Despite the promising performance of these methods in certain scenarios, several challenges persist. Firstly, generation quality heavily relies on large-scale annotated data, leading to poor generalization in few-shot scenarios. In addition, the black-box nature of the diffusion process makes topological logic difficult to trace. Furthermore, generated results often violate actual traffic rules. Future work needs to develop few-shot learning paradigms, explore topology-aware interpretability visualization techniques (such as persistence diagrams), and establish generation mechanisms constrained by physical rules.

\subsubsection{Hierarchical Structure and Progressive Reasoning}
Hierarchical progressive methods simulate human cognitive processes by decomposing complex topological problems into parallelizable subtasks through hierarchical abstraction. The core concept is "local construction-global integration", and a typical framework consists of a feature extraction layer, a topology reasoning layer, and a path optimization layer. The polyline Transformer detection network \citep{hu2023polyroad} employs vector representations instead of pixel-level masks, increasing inference speed compared to traditional methods, validating the efficiency advantages of hierarchical representation. The multi-agent hierarchical control system \citep{zhang2024ad} introduces mid-level language-driven commands to decouple high-level reasoning from low-level control. In addition, \cite{xie2025seqgrowgraph} proposes a lane graph autoregressive expansion framework, which achieves modeling of complex structures like roundabouts through incremental node addition and dynamic adjacency matrix updates.

For this kind of methods, the current bottlenecks lie in inter-level error propagation and real-time performance balancing. Autoregressive expansion models are prone to cumulative errors, and multi-agent collaboration requires solving communication latency issues. Future research should explore hierarchical error correction methods based on attention mechanisms, develop edge computing architectures to enhance real-time responsiveness, and integrate spatiotemporal hierarchical representations to improve adaptability in dynamic environments.

\subsubsection{Multimodal Fusion and Map Priors}
These methods construct robust environmental perception models by integrating heterogeneous sensor data with map prior knowledge. Its technical core lies in cross-modal attention mechanisms and a fusion architecture with prior constraints, which can compensate for performance degradation of single sensors in occluded and noisy scenarios. \cite{ye2023fusionad} proposes a multimodal fusion transformer that aligns camera semantics with LiDAR geometric features through cross-attention. \cite{yang2025deepinteraction} designs a modal interaction encoder with a dual-stream transformer architecture that preserves modality specificity while enabling directed information exchange. In \cite{pei2025sept}, authors propose a cross-modal attention topology reasoning framework that fuses standard map data with satellite imagery, achieving a topology inference F1-score of 0.87 in GPS-deprived scenarios.

Although these methods offer feasible exploration directions, they suffer from several main challenges including modal alignment accuracy, prior bias, and computational complexity. Spatiotemporal registration errors in heterogeneous data can easily lead to topological misjudgments, while the timeliness of map priors affects adaptability in dynamic scenarios. In addition, the multimodal feature processing increases hardware computational demands. Future work needs to establish dynamic update models for prior knowledge, and design lightweight fusion architectures suitable for vehicular embedded platforms.

\subsection{Summary And Limitations}
\label{sect:hd-summary}

In this section, we start from introducing the production pipeline and map specification of HD map. Through it, readers can have an overview of HD maps and have knowledge of what kind of map elements are required for autonomous driving. After, we elaborate on the following key technical points: localization, multi-trip mapping, static perception and topology generation, covering both classic and cutting-edge papers.

HD maps make autonomous driving truly possible for the first time. The industry has relied on HD maps to achieve L2 to L4 autonomous driving on highways. However, it is difficult to extend it to urban roads that are featured with more complex topology and more frequent construction or repair. This is because the high production cost, low update frequency, lack of quick response to reality change and significant data transmission of HD maps. These limitations discourage the AD industry from relying on HD maps on urban roads.

\section{Lite Map: The Conquest of Urban Roads}
By the end of 2021, most AD companies had completed commercial applications on highways, but the practice of AD on urban roads was still a challenge for them at that time. Both industry and academia started to put effort on finding an economically acceptable solution for urban roads. The concept of Lite maps started to appear and it might be most appropriate solution till today and even in the future.

\subsection{Comparison of Lite and HD Map}
\begin{figure}[thtb]
  \centering
  \includegraphics[width=1\linewidth]{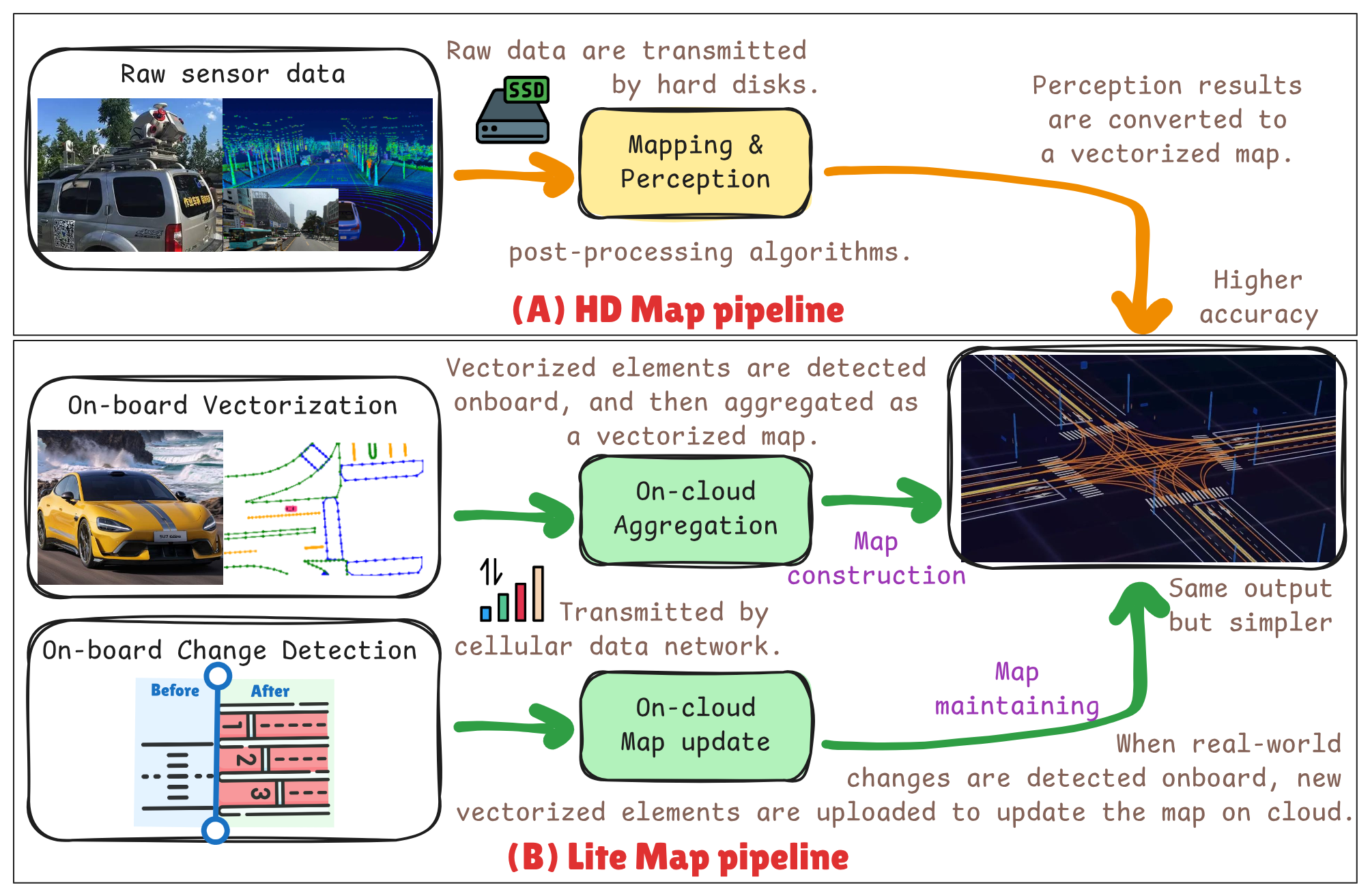}
   \caption{Comparison of Lite and HD map in terms of production pipeline and research focus.}
   \label{fig:lite-comparison}
\end{figure}

	\textit{The Lite map is an effective and acceptable solution for facilitating urban-road autonomous driving.} \figref{fig:lite-comparison} illustrates the differencesThe Lite map is an effective and acceptable solution for facilitating urban-road autonomous driving. between Lite maps and HD maps. More specifically, the differences lie in the following aspects:
	
\begin{enumerate}[label=\arabic*)]
\item
\textbf{Vehicle for map collection:} Professional surveying vehicles are used for HD map collection. These vehicles are normally owned by map providers for commercial use. In contrast, Lite maps rely on the data collected by production cars which are mass-produced cars and offered for sale to the public.  
\item
\textbf{Sensors:} Surveying-grade sensors are used for HD map collection while consumer-grade sensors are used for Lite maps. The latter are relatively much cheaper.
\item
\textbf{Production pipeline:} For HD maps, raw sensor data are transmitted directly via hard disks. Almost all mapping and perception algorithms are post-processed in a data center. In the pipeline of Lite map production, by contrast, data are collected through a crowdsourcing manner. Every car plays the role of a edge computer. The single-trip mapping and perception algorithms directly run on cars. Raw sensor data are discarded after on-board vectorization for user's privacy concern. Only vectorized elements are uploaded to a data server using cellular data network on condition that change detection reality changes are detected on board.
\item
\textbf{Map elements:} Lite maps can be regarded as a simpler version of HD maps. The map elements of Lite maps are simpler and with lower accuracy, but are sufficient for the use of autonomous driving. For example, the mapping of traffic lights and signs does not need centimeter-level accuracy at all.
\item
\textbf{Research focuses:} After inheriting the algorithms of HD maps, Lite maps are faced with new challenges. Research is focused on online vectorization, online change detection, Map differential update and traffic-flow trajectory mining. Some research points are newly encountered in the Lite map production. We will elaborate on them in this section.

\end{enumerate}

Except for the differences between Lite map and HD map, there are also many things in common, including file management, element specifications, localization and some object detection algorithms. Since these have already been discussed in \sectref{sect:hd}, we will focus on these differences in this section.

\subsection{Online Vectorization}
\subsubsection{Task Definition}
The task of online vectorization or online map learning takes as input onboard sensor observations, which normally are surrounding images and LiDAR scans, and output vectorized road elements such as lanes, boundaries, crosswalk and intersections. This task is significant in modern autonomous driving systems. It not only facilitates the Lite map construction, but also helps real-time motion prediction and planning. Although many of recent papers still refer to online vectorization as HD map learning, we still argue that it should be called Lite map due to the fact that neither its mapping accuracy nor element abundance meets the criterion of HD maps. Here we briefly divide the online vectorization, a real research hotspot in recent years, into two categories: single-frame detection and long-sequence modeling.

\subsubsection{Single-frame Detection}
The single-frame detection consumes surrounding images and/or LiDAR scans at a single time $t$, and then directly generated vectorized elements. It is independent at any timestamps and does not rely on previous or feature observations. 

HDMapNet \citep{li2022hdmapnet} proposes a novel projection module that converts features from perspective view to bird's-eye view when depth is missing. Then, a BEV decoder outputs semantic segmentation, instance embeddings and lane directions. Finally, a post-processing module is used to vectorize these predicted instances. The limitation, however, is that HDMapNet relies on the rasterized map predictions, and its heuristic post-processing step restricts the model’s scalability and performance.

The idea behind HDMapNet is segmentation followed by a post-processing DBSCAN \citep{ester1996density}. The whole framework is indirect. By contrast, VectorMapNet \citep{liu2023vectormapnet} solves the problem in an end-to-end manner. VectorMapNet is the first work designed for end-to-end vectorized map learning from onboard observations. It takes advantage of DETR \citep{carion2020end} to directly formulate the map learning as a detection task. VectorMapNet uses map queries to represent road elements in a Transformer decoder. The map queries are similar to the object queries used in DETR. The architecture of VectorMapNet comprises a BEV feature extractor, a DETR-like map element detector and a polyline generator. Since the position and shape of map elements predicted by the detector are not satisfied, the polyline generator then focuses on refining the detailed geometry of the map elements. This is actually a two-stage coarse-to-fine framework. 

Unlike VectorMapNet, MapTR \citep{liao2023maptr} proposes a single-stage framework. It designs a hierarchical query embedding scheme to flexibly encode structured map information and perform hierarchical bipartite matching for map element learning. The parallel and structured framework makes learning process higher efficiency. However, the limitation of MapTR is that it is designed to model map elements lacking physical directions, which means it does not account for map elements with practical directions such as centerlines. This will hinder the use of downstream motion prediction and planning. As an improved version, MapTRV2 \citep{liao2024maptrv2} also builds on the permutation-equivalent modeling, but does not permute centerlines and directly chooses the given order as targeted permutation. Therefore, MapTRV2 is capable of detecting centerlines with directions. 

The biggest limitation of single-frame detection is its small range of perception which is normally $60 \times 30$ meters. Worse still, in complex scenarios such as occlusions, the perception range is severely affected due to the loss of visual information.

\subsubsection{Long-sequence Modeling}
Different from the single-frame detection which only consumes images/scans at a single timestamp, the long-sequence modeling leverages historical observations in a long sequence to boost the range of perception. StreamMapNet \citep{yuan2024streammapnet} proposes a streaming strategy that encodes all historical information into the memory feature to save cost and build long-term association. This idea is borrowed from 3D objection detection works: StreamPETR \citep{wang2023exploring}, Sparse4Dv2 \citep{lin2023sparse4d} and VideoBEV \citep{han2024exploring}. More specifically, the strategy selects the former foregoing top $k$ queries based on confidence score as potential tracking queries, and then concatenates them with initialized queries. Note that the network is also DETR-like and query-based. As a result, the perception range of the local vectorized map is extended to $100 \times 50$ meters without compromising map quality.

MapTracker \citep{chen2024maptracker} argues that the aforementioned methods detect road elements anew in every frame, potentially guided by previous frames without consistency enforcement. Inspired by MeMOT \citep{cai2022memot} and MeMOTR \citep{gao2023memotr} which design memory mechanisms in tracking Transformer to preserver long-term consistency, MapTracker formulates the map learning as a tracking task using memory-based raster and vector latents for temporally consistent reconstruction. It explicitly associates tracked map elements from historical frames to further enhance temporal consistency.

MapExpert \citep{zhang2025mapexpert} raises another concern. Traditional visual detection targets are normally cube-like objects such as vehicles, pedestrians and animals that can readily represented by bounding boxes. However, the road elements in map learning task varies significantly. For instance, lanes are often smooth curves but crosswalks are usually rectangular. Such variability poses a significant challenge for DETR-like decoders. To tackle the problem, MapExpert proposes an expert-based method that addresses the challenges of representing diverse non-cubic map elements. It introduces sparse experts guided by routers, a novel auxiliary balance loss for expert load distribution, and a Learnable Weighted Moving Descent module for efficient temporal BEV fusion. MapExpert achieves state-of-the-art accuracy and efficiency on nuScenes \citep{caesar2020nuscenes} and Argoverse2 \citep{wilson2023argoverse} benchmarks.

\subsubsection{Overview And Limitations}
\begin{figure}[thtb]
  \centering
  \includegraphics[width=.7\linewidth]{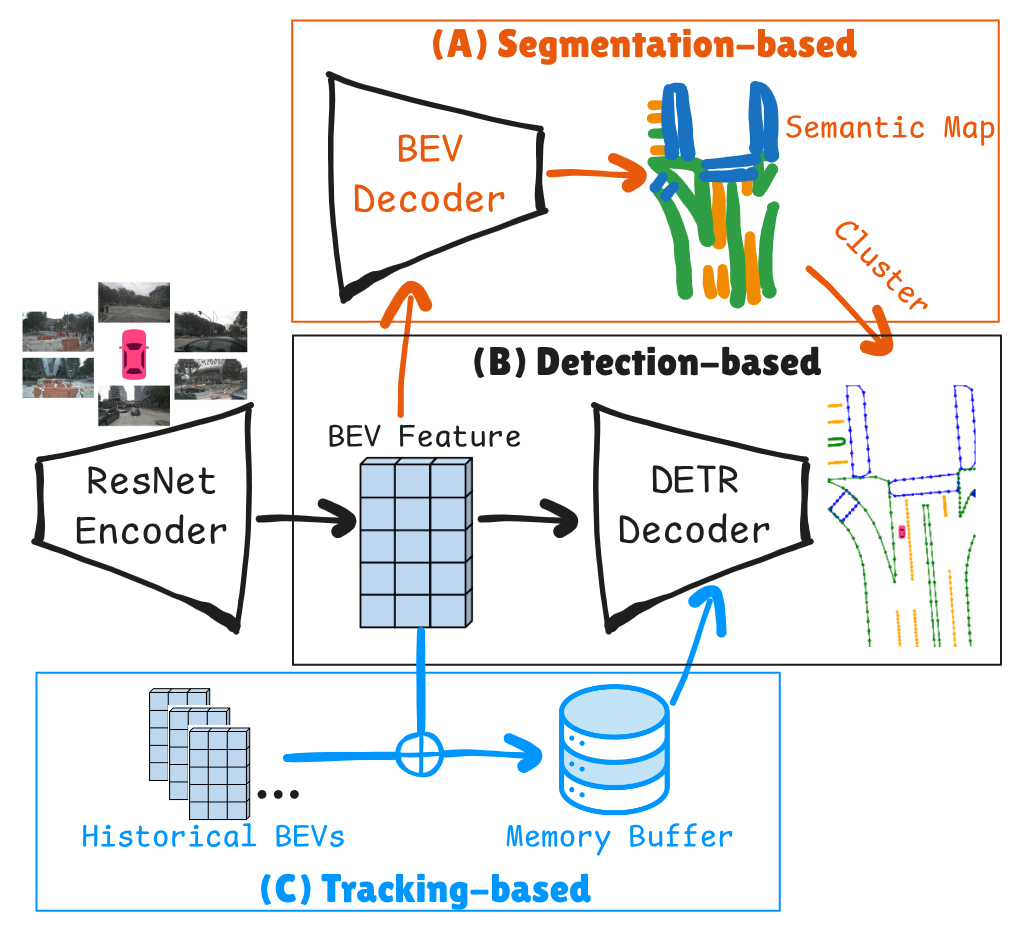}
   \caption{An architecture overview of three categories of online vectorization. (A) The segmentation-based methods generate a semantic map through a BEV decoder and then clusters it into structured elements. (B) The detection-based methods utilize a DETR-like decoder to directly detect map elements from BEV features in an end-to-end manner. (C) The tracking-based methods are featured with a memory buffer that enhances temporal consistency by aggregating historical BEV features so as to support more stable and accurate map reconstruction.}
   \label{fig:lite-online-map}
\end{figure}

Overall, the above-mentioned methods can also be summarized as segmentation-based, detection-based and tracking-based according to their formulation, as illustrated in \figref{fig:lite-online-map}. Additionally, we provide a table listing recent open-source work on online vectorization. Readers can refer to Appendix.A for quick access.

Despite the increasing popularity on the online vectorization for the online Lite map learning, the state-of-the-art performance in terms of both mapping accuracy and element abundance is still far from HD map. Many of recent works can only generate lane (centerline), boundary and crosswalk which are obviously not enough to support autonomous driving. Therefore, the online vectorization is still an open problem. We advocate the proposal of datasets with more element annotations from industry. In addition, methods regarding zero-shot, few-shot and transfer learning should be considered to enable networks to detect more types of road elements.

\subsection{Crowd-sourcing Map Maintenance}

Map maintenance strongly correlates with two tasks: change detection and map update. 

\textbf{Change detection:} The change detection aims to identify road structural changes that are misaligned with map database. We briefly classify this task into two types: post-processed change detection and online change detection. In post-processed change detection, changes are detected after data (e.g., constructed maps or images) are transmitted to a data server. On the contrary, online change detection is run onboard and only when changes are detected beforehand will data be uploaded.

\textbf{Map update:} The map update is to replace outdated map data with new one. It can also be divided into two types: direct update and incremental update. In direct update, the map for a tile or patch (we have introduced in \sectref{sect:hd-file}) is re-surveyed and will replace an old tile or patch directly. The remaining work is topological connections with adjacent tiles and patches. By contrast, the incremental  update operates on smaller unit, namely roads. It replaces a road with an updated one and fix its topology. Therefore, the incremental update is also called road-level update.

Post-processing change detection and direct update are generally used in traditional HD map maintenance. \cite{wijaya2024high} provides a well overview on them. Related works usually involve semantic map comparison \citep{jo2018simultaneous, zhang2021real} and images comparison \citep{kataoka2016semantic, kannan2024zeroscd} of the same location at two different dates. However, such maintenance frameworks would be highly impracticable for modern crowdsouring map maintenance. The reason is threefold:
\begin{enumerate}[label=\arabic*)]
\item
Volume of data transmission. Since the transmission is done by consuming users' data plan subscribed to a mobile network operator, it is impractical to send large data packages (e.g., high-resolution images or point cloud maps) from onboard systems. 

\item
Users' privacy. Images and trajectories contain much sensitive information. Uploading such data is at a risk of violating users' privacy.

\item
Map freshness. The change of road structure and topology should be quickly detected and updated to maintain map freshness.
\end{enumerate}

\begin{figure}[thtb]
  \centering
  \includegraphics[width=.7\linewidth]{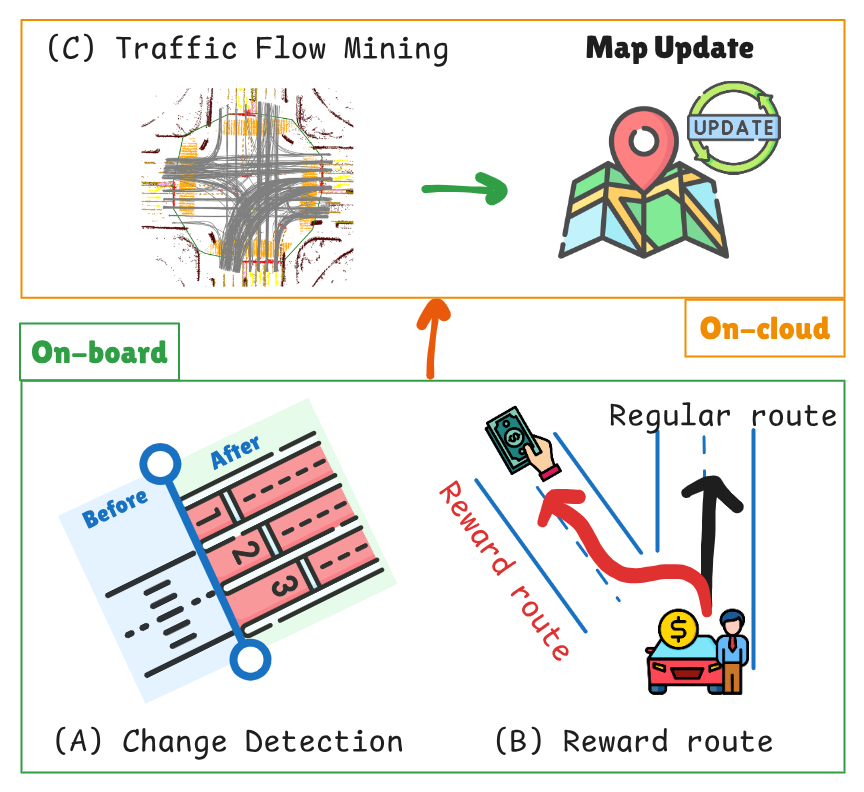}
   \caption{A modern system for dynamic map maintenance through crowd-sourced vehicle data. It is divided into on-board and on-cloud components. (A) Change Detection: identifying road structural changes (e.g., lane modifications) by comparing before-and-after states, and (B) Reward Route: drivers are incentivized to take specific routes to gather valuable data. (C) Traffic Flow Mining: extracting road elements by analyzing traffic patterns. This collaborative framework ensures that Lite map remains accurate and up to date.
}
   \label{fig:lite-maintenance}
\end{figure}

As we discussed in \sectref{sect:hd-summary} that urban roads are characterized by more frequent construction and repair compared to highways, the Lite map is hence required to be daily updated for safe autonomous driving. Reliable online map learning together with online change detection seems to be the correct solution. Taking advantage of massive crowd-sourced data, many AD companies nowadays employ a multi-source updating system for dynamic map maintenance. A typical system framework is illustrated in \figref{fig:lite-maintenance}, where three data sources contribute to map updates: change detection (\sectref{sect:lite-detect}), reward route (\sectref{sect:lite-reward}) and traffic flow mining (\sectref{sect:lite-mining}).

\subsubsection{Online Change Detection}
\label{sect:lite-detect}
The online change detection is designed to detect structural or topological road differences from the pre-downloaded map. During normal driving, data will not be uploaded to avoid excess	data transmission and privacy violation. But when a change is detected, the local Lite maps learned by online vectorization will be uploaded for map updating. Sometimes, low-resolution images will be uploaded as well for manual quality assessment.

ExelMap \citep{wild2024exelmap} rethinks map update as element-based change detection. It proposes the first end-to-end explainable element-based map change detection architecture. The architecture is based on LaneSegNet \citep{li2023lanesegnet} which consists of a BEV encoder, a Transformer decoder and a lane segment predictor. ExcelMap further adds two distinct, independently operating heads with binary output for element-wise deletion and insertion detection. Both change detection heads are simply linear layers. The final output are twofold: an updated map representation and a novel change map with a change status: unchanged, inserted or deleted.

CleanMAP \citep{shaw2025cleanmap} introduces a Multimodal Large Language Model (MLLM)-based distillation framework that assigns quantitative confidence scores (0–10) to crowdsourced  map data by evaluating lane visibility parameters (e.g., blur, lighting, occlusion). It employs a dynamic piecewise confidence-scoring function coupled with a confidence-driven local map fusion strategy to rank and fuse the most reliable local maps for updates . CleanMAP ensures that changes in local maps are accurately aligned with the confidence score-based map, maintaining consistency and accuracy in the map update.

RTMap \citep{du2025rtmap} handles road change detection as part of its real-time, on-vehicle recursive mapping system. As the vehicle navigates, RTMap continuously compares new sensor observations against the existing pre-downloaded map. It uses a probabilistic model that considers positional uncertainty to determine whether structural changes have occurred in the road environment. Detected changes are treated as loop-edge constraints within RTMap’s pose graph optimization framework. This enables the system to align submaps consistently and update the global map structure on the fly, supporting ongoing fusion of multiple agent observations to refine road element positions and update the prior HD or Lite map over time.

The open problems for online change detection are twofold:
\begin{enumerate}[label=\arabic*)]
\item
\textbf{Rate of false alarm.} According to the author's observation in the industry, most companies' online change detection algorithms have a high false alarm rate. Many change notifications uploaded to the cloud are wrong due to various reasons. These reasons may include temporary road occupation, perception errors and sensor occlusions. 

\item
\textbf{Online map repair.} In current autonomous driving practices, if a road change is detected, the AD system will call the user to take over the vehicle. This will inevitably affect the user experience. Therefore, it is a very meaningful task to repair the car-side map in time according to the change detection results. 

\end{enumerate}

These interesting yet challenging problems are real demands from AD companies. We urge the academia to pay more attention on them.

\subsubsection{Reward Route}
\label{sect:lite-reward}
For some newly built remote roads where few vehicles travel on them, it would be difficult to quickly build Lite map by spontaneous crowd-sourcing data alone. As a supplementary data source, the navigation software provides a reward route for nearby users. Users can choose to accept the reward route assignment or reject. As shown in \figref{fig:lite-maintenance}(B), if accept, users will be navigated to make a detour for data collection. In return, users will get a cash reward. Many companies, such as Uber and Huawei, have adopted such reward mechanism to improve crowd-sourcing update speed. The key research question is how to reasonably distribute reward routes and identify potential users.

\subsubsection{Traffic-flow Trajectory Mining}
\label{sect:lite-mining}
Massive users provide anonymous vehicular trajectory data. The task of traffic-flow trajectory mining is to extract useful information from these data, contributing to on-cloud Lite map maintenance. Here, we summarize the possible road elements that can be mined with some representative works as follows.

\begin{enumerate}[label=\arabic*)]
\item
\textbf{SD road network.} \cite{gao2021automatic} proposes a geospatial big data-driven framework for automatic urban road geometry extraction. The test is conducted on Didi Chuxing GAIA \citep{didi2019gaia} which is a publicly available crowd-sourcing taxis dataset. The framework consists of three main steps: trajectory compression to reduce data noise and redundancy, clustering to identify road centerlines, and vectorization to generate structured road map elements. \cite{li2024df} presents DF-DRUNet, a decoder fusion model that combines remote sensing images and GPS trajectories for road network extraction. It uses two parallel dilated Res-U-Net to effectively learn complementary features.

\item
\textbf{Intersection topology.} \cite{zhao2020automatic} proposes CITT, a three-phase framework for calibrating road intersection topology using vehicle trajectory data. It includes trajectory quality improvement, core zone detection, and topology calibration within the intersection influence zone, identifying incorrect or missing turning paths by analyzing unmatched trajectories. CITT effectively handles noisy data and diverse intersection shapes, achieving high-accuracy intersection topology updates. \cite{qin2023traffic} proposes a fully automatic and scalable framework for generating topological maps of complex intersections by leveraging crowdsourced semantic data and traffic flow information. The system infers intersection topology directly from observed vehicle movements without manual annotation.

\item
\textbf{Intersection geometry.} \cite{li2017extraction} proposes a novel intersection extraction method based on GNSS vehicle trajectories. It applies mean-shift clustering to trace road segments and estimate accurate road orientations. Dominant orientations around a location are then used to determine intersection positions, ensuring both geometric and topological correctness. \cite{tang2019novel} detects intersection boundaries via turning angle clustering, identifies entrances and exits using a novel approach, and reconstructs geometric structures with assigned turning rules. \cite{chen2020extended} enhances intersection detection from noisy, low-sampling GNSS data by proposing a three-step method. The steps include (a) stay-point detection: to filter out false turn-points unrelated to intersections, improving the reliability of turning data; (b) turn-point compensation: to estimate accurate turning locations by analyzing geometric patterns and hotspot areas, using a density-based turning angle model to refine sparse trajectory data; and (c) post-classification: to refine intersection candidates by clustering and classification using geometric features.

\item
\textbf{Interchange networks.} \cite{jiao2025automatic} introduces a novel method for generating accurate road interchange networks using a forward and reverse tracking mechanism based on long-term trajectory continuity. The process includes (a) preprocessing: to filter out non-interchange and abnormal trajectories; (b) subgraph extraction: to detect potential transition nodes (where roads diverge or converge); (c) forward and reverse Tracking: to identify divergence nodes through forward tracking and convergence nodes; and (d) two-stage fusion: to merge forward and reverse tracking results.

\item
\textbf{Centerline.} \cite{chen2021automatically} extracts centerlines from low-frequency GPS trajectory data using a divide-and-conquer, intersection-first approach. This approach effectively balances precision and recall under low-sampling conditions, producing smoother and more connected road centerlines than traditional KDE-based approaches. \cite{dal2022road} employs morphological analysis to generate elongated polygons that represent road ribbons, followed by skeletonization to reconstruct the road centerlines and the overall road network. The method is based on raw trajectory polylines, avoiding raster resampling. 

\item
\textbf{Road boundary.} \cite{yang2018method} proposes a road boundary extraction approach using Delaunay triangulation (DT) on crowdsourced GPS vehicle traces. It constructs DT and Voronoi diagrams over the interpolated trajectories to derive geometric descriptors (e.g., Voronoi cell area, triangle edge length). Then, a boundary detection model followed by a region-growing algorithm is applied to extract complete road boundaries.

\item
\textbf{Lane} \cite{arman2021lane} proposes a three-step automatic method for constructing lane-level routable digital maps from trajectory data. The steps include a QuickBundles clustering for road node detection, a Fréchet distance-based dissimilarity matrix, and a Gaussian Mixture Model (GMM) for estimating lane positions as well as lane numbers.

\end{enumerate}

There are many other applications of traffic-flow data such as crash rate analysis, real-time traffic dispatching and travel time estimation (ETA). They will not be discussed further in this paper because we are more focused on extracting static map elements using traffic-flow.

\subsection{Summary And Limitations}
In this section, we first compare the similarities and differences between HD maps and Lite maps, highlighting the new challenges encountered during the Lite map stage. Then, we conduct an in-depth discussion on the research hotspots in the Lite map, including online vectorization, online change detection and traffic-flow trajectory mining. Through this section, readers should be able to understand why Lite maps are more advanced than HD maps.

Lite maps have helped autonomous driving companies expand from highways to urban areas, realizing the true commercialization of self-driving cars. The explicit map representations connect modular tasks (i.e., perception, prediction and planning), but they also make AD systems suffer from accumulative modular errors and deficient task coordination. This makes us to think whether the Lite map is the ultimate form for AD systems? The answer is still unclear now, but researchers are working on more human-like end-to-end AD systems where maps are implicitly represented in neural networks. 

\section{Implicit Map: March Towards End-to-End AD}

\begin{figure}[thtb]
	\centering
  	\includegraphics[width=1\linewidth]{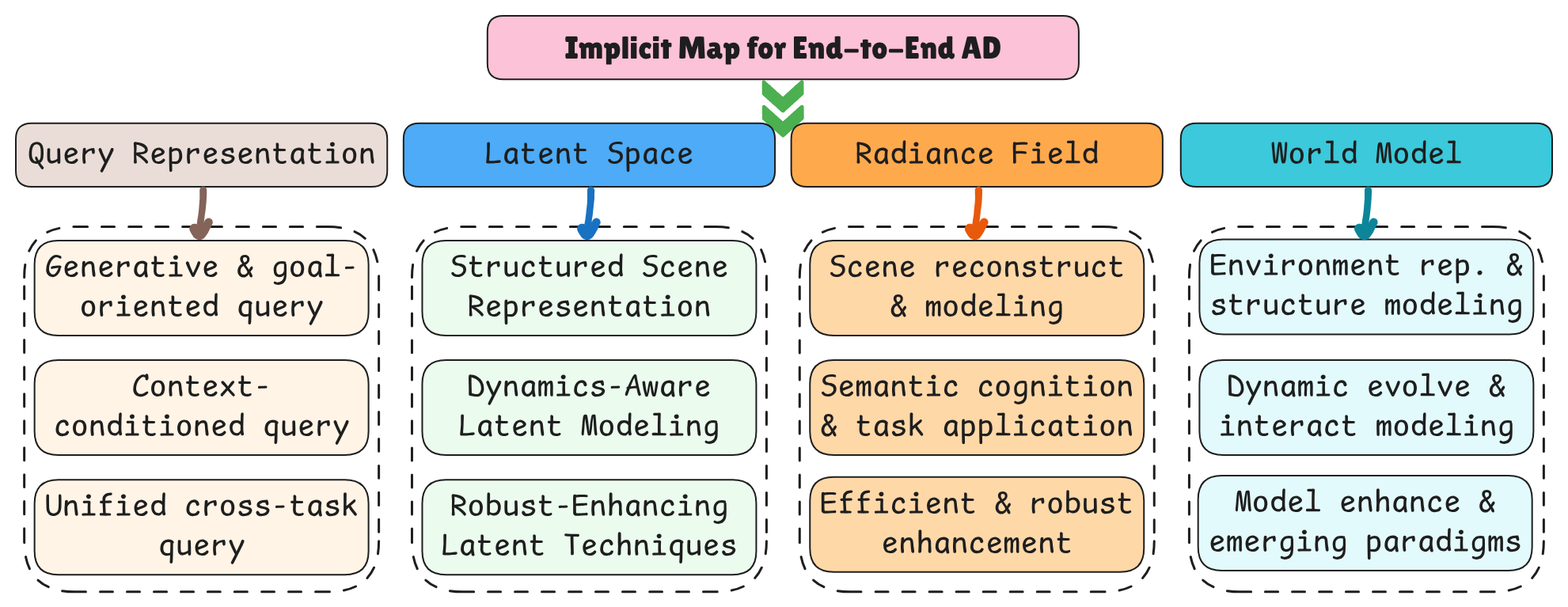}
  	\caption{Implicit map-based methods for autonomous driving}
  	\label{fig:imp-overview}
\end{figure}

\begin{figure}[thtb]
	\centering
  	\includegraphics[width=1\linewidth]{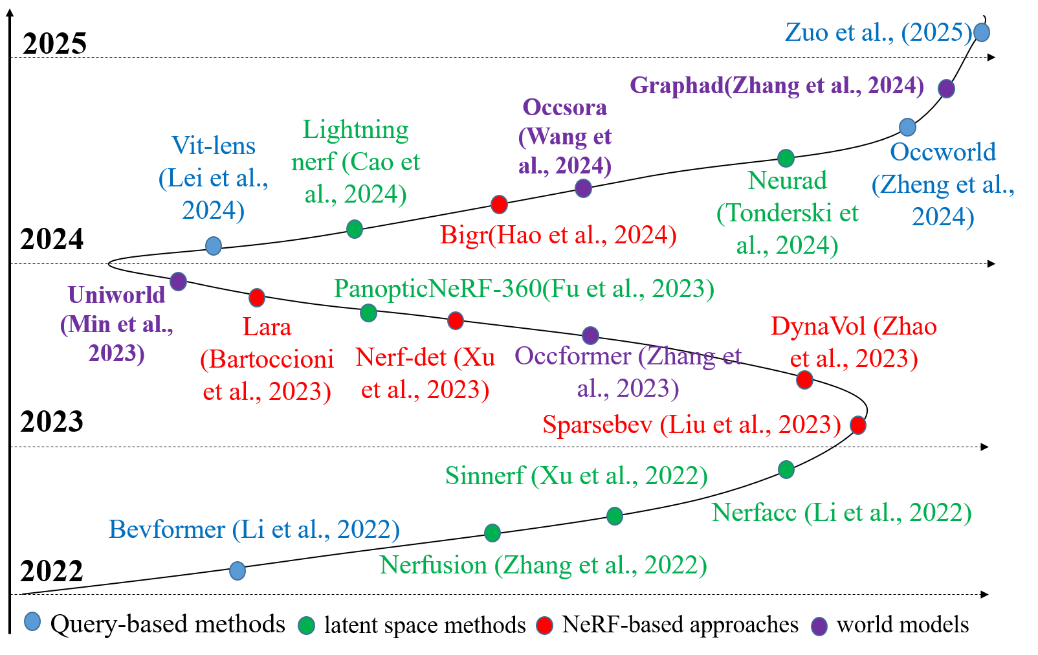}
  	\caption{Chronological overview of several representative implicit map methods}
  	\label{fig:imp-chrono}
\end{figure}

As shown in \figref{fig:imp-overview}, this section aims to systematically review the Implicit map in autonomous driving, structured into four kinds of technologies. First, query-based representations are examined through three paradigms: generative goal-directed synthesis, context-conditioned formulation, and unified cross-task architectures. Subsequently, latent space methods integrate spatial structuring and temporal dynamics to compress high-dimensional sensory data into structured latent spaces. Next, Neural Radiance Fields (NeRF)-based approaches focus on 3D scene reconstruction, semantic understanding, and efficiency optimization techniques. Finally, world models encompass environmental representation, dynamic evolution modeling, and model enhancement technologies for predictive simulation. Subsequent sections will critically analyze strengths, limitations, and trade-offs including efficiency-expressiveness balance and safety verification to guide future research directions. \figref{fig:imp-chrono} provides the chronological overview of several representative implicit map methods.

\subsection{Query Representation Methods}
\begin{figure}[thtb]
	\centering
  	\includegraphics[width=.7\linewidth]{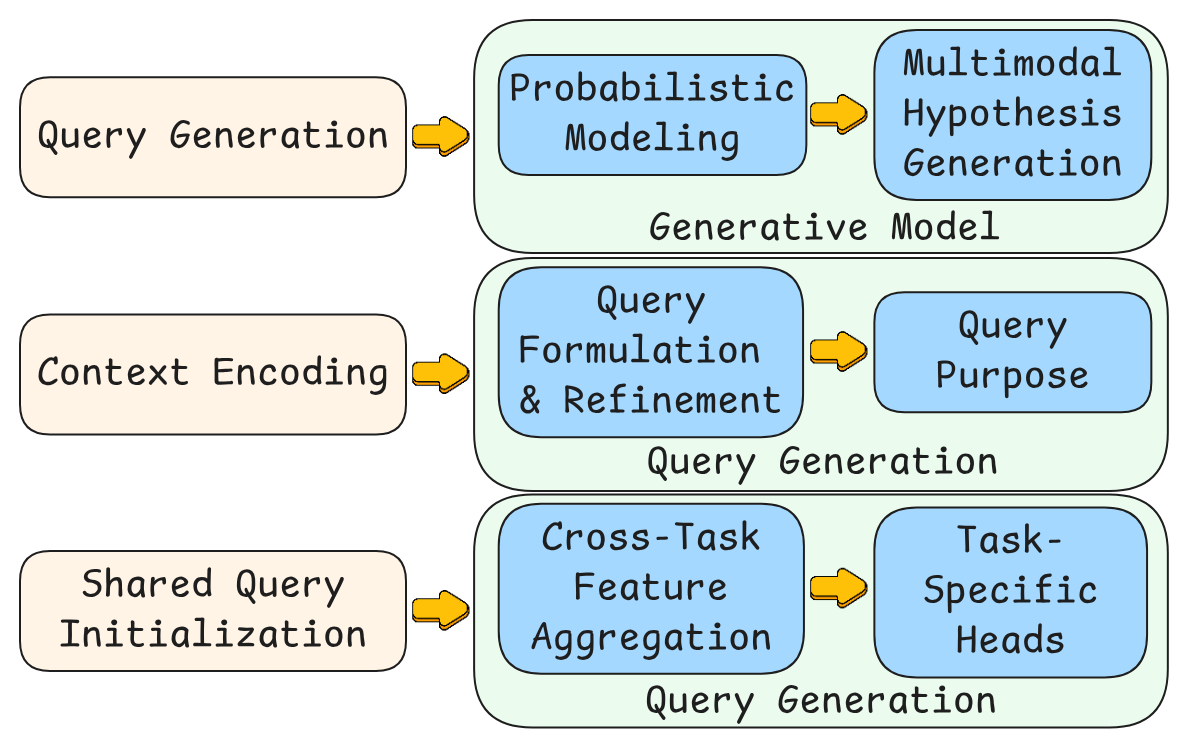}
  	\caption{Three methodological lenses of query-based representations}
  	\label{fig:imp-query}
\end{figure}

Query-based representations have revolutionized end-to-end autonomous driving by enabling structured interaction between perception, prediction, and planning modules. As shown in \figref{fig:imp-query}, existing works can be divided into three methodological lenses: generative and goal-oriented query synthesis, context-conditioned query formulation, and unified cross-task query architectures. We critically examine how these approaches balance expressiveness, efficiency, and interpretability while identifying fundamental limitations in temporal coherence, causal reasoning, and safety verification that impede real-world deployment.

\subsubsection{Generative \& Goal-Oriented Query Synthesis}
This paradigm leverages goal-directed mechanisms to produce diverse, multimodal trajectory hypotheses. Potential-based diffusion motion planning \citep{luo2024potential} pioneers diffusion-based query generation for probabilistic trajectory sampling, capturing complex maneuver distributions through composable potential functions. Trajectory Prediction using LSTM with Attention mechanism \citep{ahmadi2023human} injects intent awareness through goal-conditioned queries, enabling human-like anticipation of agent objectives by weighting historical trajectory features. GaussianWorld \citep{zuo2025gaussianworld} extends continuity by modeling trajectories as queryable spatiotemporal Gaussian fields, while Tracking dynamic flow \citep{tian2024tracking} integrates optical flow priors to anchor predictions in observed motion dynamics. Point Cloud Forecasting as a Proxy for 4D Occupancy Forecasting \citep{khurana2023point} further generalizes this concept to 4D occupancy spaces by rendering point clouds from predicted occupancy.

These methods demonstrate their distinct strengths. The robust probabilistic modeling capability enables the capture of multimodal trajectory distributions in complex scenarios. Besides, the generative framework inherently supports uncertainty quantification, providing a theoretical basis for risk assessment. Despite their representational power, these methods exhibit several fundamental limitations:
\begin{enumerate}[label=\arabic*)]
\item
The trade-off between the computational efficiency and representational power: Probabilistic modeling (e.g., diffusion processes) requires hundreds of iterative refinements, conflicting with real-time planning constraints.

\item
Input quality dependency: Goal-conditioned approaches assume perfect intent recognition and environmental perception, which degrade under noisy or ambiguous inputs.

\item
Safety mechanism deficiency: None of the generative frameworks incorporate formal verification of safety boundaries, risking physically invalid trajectory proposals.
\end{enumerate}

\subsubsection{Context-Conditioned Query Formulation}
These techniques contextualize queries using linguistic, temporal, or structural cues to align with human cognition and environmental constraints. DAGGER-Q \citep{mullins2018accelerated} early combined query mechanisms with imitation learning, using confidence and risk queries to dynamically trigger expert interventions. This reduced data annotation costs and provided an early framework for context-driven efficient learning. A Language Agent for Autonomous Driving \citep{mao2023language} and OccWorld \citep{zheng2023occworld} fuse natural language commands with motion/occupancy queries for human-aligned planning, while QA-Drive \citep{khan2023exploring} uses question-answer pairs to generate explainable planner justifications. DPBridge \citep{ji2024dpbridge} ensures temporal consistency through decoupled past/future queries, and IC-Mapper \citep{zhu2024ic} encodes road topology priors for map-consistent perception. Think-on-Graph \citep{sun2023think} introduces traffic rule verification via knowledge graph queries, and MAPO \citep{chen2024mapo} employs LLMs to abstract high-level decisions into symbolic queries.

These approaches naturally integrate human instructions and common-sense knowledge, enhancing the intuitiveness of human-machine interaction, and dynamically adjusting query strategies to adapt to scene changes, improving robustness in complex environments. However, they suffer from several core challenges:
\begin{enumerate}[label=\arabic*)]
\item
Ambiguity in cross-modal grounding: Natural language commands often contain spatial vagueness (e.g., "near the tall building") which resists precise geometric interpretation.

\item
Causal explainability gap: QA-based justification frameworks correlate with planning decisions but fail to establish causal relationships between query features and output actions.

\item
Temporal robustness limitations: Historical context accumulation introduces drift in long-horizon planning, particularly in dynamic environments with occluded agents.
\end{enumerate}

\subsubsection{Unified Cross-Task Query Architectures}
This category unifies disparate driving tasks through shared query representations that bridge perception-planning gaps. Neural Motion Planner with Query-based Occupancy \citep{dalal2024neuralmp} first proposed planning-driven sparse spatiotemporal queries, enabling direct perception-planning interaction. BEVFormer \citep{li2022bevformer} efficiently aggregated spatiotemporal features via dynamic BEV queries, with its sparse query mechanism widely adopted by subsequent frameworks like UniAD \citep{hu2023planning}. MetaQuery \citep{wang2022unified} enables few-shot adaptation to novel scenarios through learnable query initialization. 

Unified cross-task query architectures break down traditional modular barriers, enabling end-to-end optimization from perception to planning. They efficiently share multimodal features, significantly reducing information redundancy; and enhance complex task coordination capabilities, supporting multi-objective optimization in dynamic scenarios. However, they encounter multi-task optimization conflicts, where perception accuracy and planning safety objectives compete in gradient descent. Additionally, they face the exponential growth in system complexity with task scale, failing to meet scalability requirements for urban-level scenarios.

\subsubsection{Summary}
Generative and goal-oriented methods capture multimodal trajectory distributions through probabilistic modeling, providing uncertainty quantification for complex scenario decision-making, yet face inherent conflicts between computational efficiency and real-time performance. Context-conditioned approaches achieve deep integration of human intent and environmental features, enhancing system adaptability and explainability, while struggling with contextual ambiguity resolution and causal reasoning bottlenecks. Unified cross-task architectures break traditional modular barriers, enabling end-to-end perception-planning optimization via shared latent spaces, but are constrained by multi-objective conflicts and safety verification complexity. These three categories collectively advance query-driven autonomous systems, yet none have fully resolved the robustness-efficiency-safety trilemma in dynamic open environments.

\subsection{Latent Space Representation Methods}
\begin{figure}[thtb]
	\centering
  	\includegraphics[width=.75\linewidth]{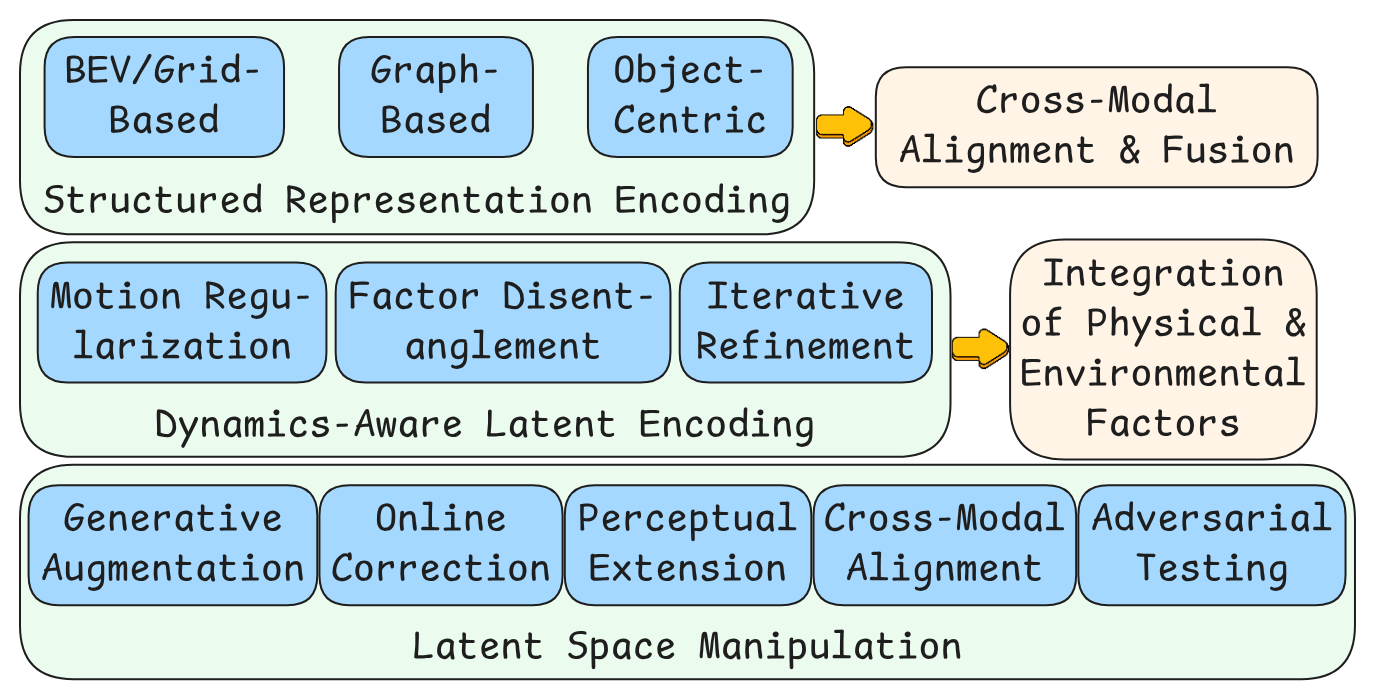}
  	\caption{Three methodological lenses of latent space representation methods}
  	\label{fig:imp-latent}
\end{figure}

Latent space representations have emerged as a critical enabler for end-to-end autonomous driving systems, compressing high-dimensional sensor data into compact, structured feature spaces for efficient perception, prediction, and planning. As shown in \figref{fig:imp-latent}, we categorize the existing methods into three core methodological paradigms: Structured Scene Representation, Dynamics-Aware Latent Modeling, and Robustness-Enhancing Latent Techniques.

\subsubsection{Structured Scene Representation}
This paradigm focuses on geometrically consistent and topologically meaningful latent organization of driving scenes. SparseBEV \citep{liu2023sparsebev} optimizes efficiency through sparsified convolutional latent grids, addressing computational bottlenecks in real-time applications with its fully sparse architecture for panoptic occupancy prediction. Object-centric organization is pioneered by DynaVol \citep{zhao2023dynavol}, which allocates dedicated latent slots to dynamic objects via 3D voxelization, and ViT-Lens \citep{lei2024vitlens}, which employs hierarchical transformers for multi-task feature sharing across omni-modal inputs. Cross-modal alignment is achieved in LaRa \citep{bartoccioni2023lara} through radar-augmented BEV fusion and COLA \citep{saeed2021contrastive} via contrastive learning for sensor-invariant representations.

These methods excel in efficient spatial information compression by transforming high-dimensional sensor data into compact latent spaces through techniques such as BEV projection and sparse convolutional grids, enabling real-time scene perception as demonstrated in BEVFormer and SparseBEV. Furthermore, they exhibit strong multi-modal fusion capabilities, facilitating cross-sensor alignment (e.g., radar-augmented BEV fusion in LaRa and contrastive learning for sensor-invariant representations in COLA). However, they face inherent limitations: BEV projections suffer from depth estimation errors and occlusion artifacts, leading to feature degradation in distant or occluded regions; graph-based approaches like VectorMapNet rely on predefined topological templates, limiting adaptability to novel road layouts (e.g., temporary construction zones); and object-centric methods depend heavily on upstream detection accuracy, resulting in cascading errors for small or ambiguous objects.

\subsubsection{Dynamics-Aware Latent Modeling}
These techniques embed temporal dynamics and motion priors directly into latent spaces for accurate trajectory forecasting and planning. SceneFlow \citep{xiang2023self} integrates optical flow as motion regularization, enabling smooth latent space transitions across frames through self-supervised learning of rigid flow consistency. DriveLatent \citep{zhu2019learning} disentangles static and dynamic scene factors via a variational autoencoder, allowing separate manipulation of environmental elements and agent motions. LatentDiffuser \citep{li2023efficient} leverages diffusion models within low-dimensional latent spaces to refine planning trajectories iteratively, bridging perception and control with energy-guided sampling. Physical environment interactions are encoded in Multi-Body Neural Scene Flow \citep{vidanapathirana2023multi}, which embeds terrain roughness features for off-road planning via isometry-regularized flow estimation, and HADNet \citep{shan2023hybrid}, which uses holistic attention to fuse agent-road-traffic light dynamics.

These methods effectively embed temporal dynamics into latent spaces through different techniques, enhancing trajectory prediction accuracy for complex multi-agent interactions. They support multimodal future trajectory generation via diffusion models (LatentDiffuser) and adaptive motion priors, accommodating diverse traffic scenarios. Additionally, these methods integrate physical environmental cues such as terrain roughness and road-traffic light dynamics, improving adaptability to unstructured environments. Nevertheless, critical challenges persist. Firstly, computational complexity remains high due to iterative optimization in diffusion-based planners and multi-body motion decomposition, hindering real-time performance for high-speed driving. Secondly, nonlinear motion modeling (e.g., emergency maneuvers) remains inadequate, leading to flow field distortions in extreme scenarios. Finally, vulnerability to data sparsity limits generalization to rare events like accidents or sudden obstacles, increasing overfitting risks.

\subsubsection{Robustness-Enhancing Latent Techniques}
This category addresses domain adaptation, sensor degradation, and scenario generalization challenges through latent space manipulation. GELA \citep{tian2024latent} employs diffusion-based latent augmentation to synthesize rare scenarios, enhancing robustness to long-tail events via conditional diffusion models on dynamic graphs. LAC \citep{hao2024bigr} corrects calibration errors through online affine transformations in binary latent codes, maintaining performance under sensor misalignment. \cite{mahara2024generative} generates latent maps beyond sensor range using contrastive learning-equipped GANs, extending perceptual horizons with 14.6\% higher MAP than traditional methods. Adversarial robustness is addressed by GANav \citep{shukla2023generating}, which tests planners against latent-space perturbations, while \cite{jia2021scaling} aligns visual and linguistic latent spaces for improved command interpretation via contrastive language-image pre-training.

These methods strengthen model generalization through generative data augmentation (e.g., GELA’s diffusion-based synthesis) and adversarial training (e.g., GANav’s latent space perturbations), addressing long-tail events such as adverse weather and sensor failures. They achieve cross-domain adaptability via contrastive learning (COLA) and vision-language alignment (CLIP-Drive), enabling sensor-invariant and scene-agnostic representations. Computational efficiency is optimized through binary latent coding (EcoLane) and sparse feature enhancement (LAC), making them suitable for onboard embedded systems. However, these techniques exhibit notable limitations: 1) Generative augmentation heavily depends on high-quality real-world data distributions, risking physically implausible samples in data-scarce scenarios (e.g., rural roads); 2) adversarial training and domain adaptation lack quantifiable robustness boundaries, leading to performance degradation under unseen sensor degradation modes (e.g., lens contamination); and 3) safety validation mechanisms for augmented samples are absent, potentially introducing misleading features such as unsafe driving patterns in extreme scenario generation.

\subsubsection{Summary}
These methods advance latent space representation from spatial structuring, temporal dynamics, and robustness perspectives, respectively, yet all face fundamental trade-offs between physical consistency and computational efficiency. Future research should prioritize integrating geometric-dynamic joint modeling to mitigate cascading errors from BEV projection and motion estimation, embedding physical constraints into generative augmentation to ensure safety and realism of synthetic samples, and developing lightweight architectures to balance model complexity with real-time inference requirements for autonomous driving systems.

\subsection{Neural Radiance Fields based methods}
\begin{figure}[thtb]
	\centering
  	\includegraphics[width=.85\linewidth]{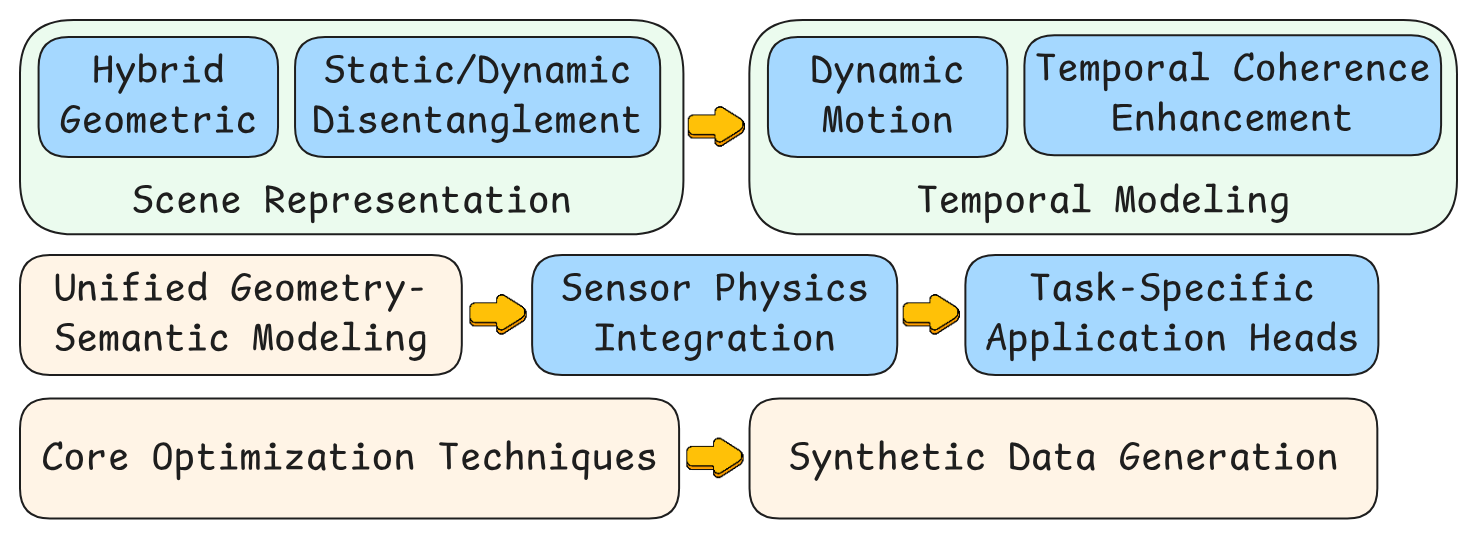}
  	\caption{Three methodological lenses of neural radiance fields-based methods}
  	\label{fig:imp-radiance}
\end{figure}

With the groundbreaking progress of Neural Radiance Fields (NeRF) in 3D scene representation, their application in autonomous driving is undergoing a critical transition from theoretical exploration to system integration. As shown in \figref{fig:imp-radiance}, we categorize the existing works into three kinds: the scene reconstruction and modeling methods, semantic understanding and task applications methods, and efficiency and robustness enhancements methods.

\subsubsection{Scene Reconstruction and Modeling}
Recent advancements in Neural Radiance Fields (NeRF) for scene reconstruction have focused on balancing accuracy, speed, and dynamic environment handling. Hybrid approaches like Street Gaussians \citep{yan2024street} combine 3D Gaussians with point clouds to achieve 135 fps rendering for dynamic urban scenes, prioritizing real-time performance through explicit geometric representations. In contrast, S-NeRF++ \citep{chen2025snerf} employs a dual-branch architecture to disentangle static backgrounds and dynamic vehicles, improving reconstruction fidelity but increasing computational complexity. Dynamic modeling techniques such as Neural Radiance Flow \citep{du2021neural} introduce differentiable scene flow for unsupervised motion estimation, while TimeNeRF \citep{hung2024timenerf} enhances temporal coherence through recurrent rendering. Occlusion handling solutions include GANeRF \citep{roessle2023ganerf} with adversarial completion, reducing LPIPS by 50\% on Tanks and Temples. Lightning-NeRF \citep{cao2024lightning} integrates LiDAR for illumination decomposition, achieving 5× faster training through hybrid representation.

These methods demonstrate notable strengths in balancing efficiency and fidelity, as hybrid approaches like Lightning-NeRF and explicit representations such as Street Gaussians achieve 5-10× speedups while maintaining state-of-the-art reconstruction quality. These techniques excel in dynamic handling through dual-branch architectures like S-NeRF++ and 4D radiance flows such as Neural Radiance Flow, enabling robust modeling of moving objects and temporal changes. Despite these advancements, several limitations persist, including artifacts from transient dynamics where heuristic handling of sudden motion (e.g., pedestrians, fast vehicles) causes rendering distortions, as observed in EmerNeRF's limited improvement for extreme transience. In addition, View-dependent ray sampling strategies remain a bottleneck, limiting real-time performance in open-road scenarios with rapid viewpoint changes. Furthermore, illumination decomposition methods exhibit LiDAR dependency, degrading performance in LiDAR-sparse regions and restricting applicability in sensor-limited setups.

\subsubsection{Semantic Understanding and Task Applications}
Semantic NeRF advancements enable unified geometry-semantic modeling and task-specific optimizations for autonomous driving. PanopticNeRF-360 \citep{fu2023panopticnerf} generates panoramic urban scene annotations with view-consistent panoptic labels, reducing manual labeling costs by 1.5 hours per scene. GSNeRF \citep{chou2024gsnerf} fuses semantic and geometric features through a two-stage pipeline, and improves 3D scene understanding via depth-guided rendering. For 3D detection, NeRF-Det \citep{xu2023nerf} avoids point cloud conversion losses. It obtains a significant improvement on ScanNet by leveraging geometric-aware volumetric representations. NeRF-Loc \citep{sun2023nerfloc} achieves cm-level localization precision through Transformer-based feature alignment, outperforming traditional methods on Cambridge Landmarks. NeuRAD \citep{tonderski2024neurad} supports closed-loop simulation by modeling sensor physics (rolling shutter, LiDAR beam divergence), enabling safety-critical scenario testing.

Semantic understanding methods offer unified representation capabilities, as seen in PanopticNeRF-360 and GSNeRF, which enable simultaneous geometry-semantic modeling and improve label propagation accuracy. Task-specific optimizations deliver significant gains, with NeRF-Det eliminating point cloud conversion losses and NeRF-Loc achieving sub-centimeter localization precision critical for autonomous navigation. However, query latency remains problematic, with NeRF-Det and NeRF-Loc requiring 30-50ms per query, exceeding the real-time threshold for autonomous driving. Besides, physical inconsistencies also persist, as EditNeRF's latent manipulation overlooks light transport physics, causing unrealistic shadows and reflections in edited scenes.

\subsubsection{Efficiency and Robustness Enhancements}
These methods aim to address deployment barriers for real-world autonomous systems. NeRFAcc \citep{li2023nerfacc} achieves 1.5-20× training speedup via hash encoding and CUDA fusion, enabling real-time rendering. For environmental robustness, ClimateNeRF \citep{li2023climatenerf} employs degradation-aware ray sampling for rain/snow conditions, while PC-NeRF \citep{hu2024pc} uses sparse LiDAR regularization to improve reconstruction accuracy under 67\% data loss. Sensor fusion methods like NeRFusion \citep{zhang2022nerfusion} integrate Camera-LiDAR-radar via cross-attention. Additionally, SinnNeRF \citep{xu2022sinnerf} generates annotated synthetic data to mitigate real-data scarcity.

These efficiency enhancement methods deliver substantial computational gains, with NeRFAcc and Lightning-NeRF achieving 10× speedups that make real-time operation feasible for autonomous driving platforms. Environmental adaptability is strengthened through methods like ClimateNeRF and PC-NeRF, which maintain performance in adverse conditions and sensor data loss, showing an improvement in rainy scenes. Despite these advancements, they suffer several deployment challenges. Calibration sensitivity is a key issue. When sensor synchronization delays exceeding 10ms, it will cause fusion failures and degrade radar-camera consistency. In addition, the sim-real gap also persists, as material property discrepancies in SimNeRF-generated data lead to 10\% lower detection accuracy when transferred to real-world scenarios.

\subsubsection{Summary}
Neural Radiance Fields (NeRF) have transitioned from theoretical models to practical autonomous driving components, with advancements in dynamic modeling, semantic integration, and efficiency. Key achievements include 10× speedups, cm-level localization, and multi-sensor fusion robustness. However, critical gaps remain. Transient dynamics handling causes artifacts, while semantic resolution limits small object recognition, and calibration errors hinder sensor fusion. Future research should focus on spatiotemporal-consistent joint modeling, task-driven differentiable rendering, and neuro-symbolic integration to reduce annotation reliance. Addressing these barriers may enable NeRF to serve as the foundation for next-generation "environmental universal models" in autonomous driving.

\subsection{World Model}
World models constitute a critical component in autonomous driving systems, enabling the prediction of dynamic environment evolution to inform safe planning and decision-making. We categorize the existing methods into three primary kinds based on their core objectives and technical approaches: Environment Representation \& Structured Modeling, Dynamic Evolution \& Interaction Modeling, and Model Enhancement \& Emerging Paradigms.

%\begin{figure}[thtb]
%	\centering
%  	\includegraphics[width=.6\linewidth]{img/imp-worldmodel.png}
%  	\caption{Three methodological lenses of world models}
%  	\label{fig:imp-worldmodel}
%\end{figure}

\subsubsection{Environment Representation and Structured Modeling Methods}
This paradigm focuses on developing efficient, robust, and structured representations of complex driving scenes, forming the foundational state upon which predictions are made. OccFormer \citep{zhang2023occformer} exemplifies advancements in dense scene understanding by proposing an instance-aware 4D occupancy prediction framework that unifies geometric structure and dynamic evolution through agent-scene interaction. Similarly, GAIA-1 \citep{hu2023gaia} adopts an object-centric perspective, explicitly modeling key scene entities (e.g., vehicles, pedestrians) with integrated geometric constraints, aligning more closely with human scene cognition. GraphAD \citep{zhang2024graphad} abstract scene elements as nodes within a graph structure, utilizing Graph Neural Networks to learn complex spatial and interactive dependencies, thereby providing a powerful tool for relational reasoning. QuAD \citep{biswas2024quad} enhances practical utility by designing a world model capable of on-demand queries (e.g., collision checking), significantly improving planning efficiency. 

Despite these significant strides in scene comprehension and structured expression, substantial challenges persist. Dense representations like occupancy grids incur prohibitive computational and memory overhead, hindering real-time deployment, while object-centric approaches are critically dependent on the accuracy of upstream perception modules (detection/tracking), making them vulnerable to error propagation. In addition, the effectiveness of graph-based models is contingent upon the completeness and appropriateness of the defined relationships between nodes. Furthermore, vision-only models exhibit limited robustness under adverse conditions such as severe weather, heavy occlusion, or extreme lighting variations.

\subsubsection{Dynamic Evolution and Interaction Modeling Methods}
This category focuses on forecasting the future trajectory of scene states, emphasizing physical laws, multi-agent interactions, and the inherent uncertainty of predictions. Physics-based modeling is prominently featured in UniWorld \citep{min2023uniworld}, which employs coupled differential equations to jointly model agent kinematics and environmental physical effects (e.g., friction, aerodynamics), demonstrating deep integration of physics and agent behavior. \cite{fan2024perception} addresses the specific demands of off-road driving by incorporating soil mechanics into its dynamic predictions. For modeling interactions, InterSim \citep{sun2022intersim} introduces game-theoretic frameworks to explicitly capture strategic interactions (e.g., competition, cooperation) among multiple agents (vehicles, pedestrians), providing a principled approach to complex behavioral prediction. \cite{ahmadi2024curb} directly predicts joint trajectory sets for all interacting agents, mitigating the cascading errors inherent in sequential "predict-then-react" pipelines. Generative approaches are represented by DriveDreamer \citep{wang2023drivedreamer}, which harnesses diffusion models to synthesize high-quality future scene states, and OccSora \citep{wang2024occsora}, which utilizes variational auto-encoders and latent space regularization to encourage diverse predictions and alleviate mode collapse. Recognizing the criticality of uncertainty for safety, \cite{li2024bayesian} implements Bayesian Neural Networks to propagate uncertainty through the prediction horizon, forming a foundation for risk-aware decision-making. \cite{hu2024gitsr} enforces bidirectional temporal consistency losses to ensure the coherence of long-horizon predictions, indirectly enhancing prediction reliability.

Nevertheless, significant limitations remain. Physics-based models often entail high computational costs incompatible with real-time onboard requirements. Game-theoretic approaches suffer from high computational complexity and often employ simplified models of agent intent and rationality. Generative models (e.g., diffusion, VAE) are hampered by slow inference speeds, and ensuring the physical plausibility and controllability of generated outcomes is challenging. Bayesian methods for uncertainty estimation are computationally intensive and struggle to reliably disentangle epistemic (model) uncertainty from aleatoric (data) uncertainty.

\subsubsection{Model Enhancement and Emerging Paradigm Methods}
This final category encompasses efforts to augment specific critical capabilities of world models or integrate disruptive novel AI paradigms to address complex challenges and enhance applicability. \cite{greve2024collaborative} pioneers the integration of formal verification techniques to provide mathematical guarantees on prediction safety boundaries, representing a crucial step towards certifiably safe systems. \cite{seo2024metaverse} employs meta-learning strategies to enable rapid few-shot adaptation of world models to entirely new urban layouts, significantly boosting generalization and deployment efficiency. Explorations of emerging paradigms feature DriveGPT4 \citep{xu2024drivegpt4}, which innovatively utilizes Large Language Models (LLMs) to capture state transition rules and semantic scene knowledge, leveraging the powerful reasoning and knowledge representation capabilities of LLMs for complex scenario understanding. \cite{zhang2024changes} addresses data scarcity and long-tail challenges by focusing on generating high-fidelity synthetic scenarios for effective world model pre-training.

While these approaches represent cutting-edge explorations, they face distinct shortcomings. The scope of formal verification in safety-focused models is often limited and operates at high levels of abstraction, struggling to encompass the full complexity of real-world driving. Training differentiable models like NeuralDriving can be complex and prone to instability. The efficacy of meta-learning approaches is heavily dependent on the diversity of meta-training data and the inherent generalization capability of the base model. LLM-based models contend with inherent challenges, including significant inference latency, real-time performance limitations, potential "hallucinations," and difficulties in effectively grounding domain-specific driving knowledge. In addition, synthetic data generation grapples with the persistent "reality gap" between synthetic and real-world data, and ensuring the generated data distribution effectively covers critical scenarios and variations is challenging.

\subsubsection{Summary}
The world model increasingly integrates physical laws and interactive priors to improve prediction realism, while advancing structured representations (e.g., graph-based, object-centric). It balances generative models (e.g., diffusion) for diverse prediction and probabilistic techniques for uncertainty quantification. Recent research also focuses on verification, causal reasoning, and adaptability. Key challenges include computational efficiency, complex interaction modeling, robust uncertainty estimation, sim-to-real transfer, multi-modal sensor fusion, interpretability, and generalization. Addressing these requires interdisciplinary collaboration and foundation models to achieve verifiable and scalable autonomous driving.

\subsection{Conclusion for implicit map}
Collectively, these implicit map-based methods establish the theoretical foundation for modern autonomous driving representation learning, reflecting the field's evolution from modular perception systems to integrated end-to-end architectures. Cross-cutting trends include the convergence of computer vision and robotics principles, the increasing integration of physical constraints into data-driven models, and the shift toward unified frameworks that bridge perception, prediction, and planning. Fundamental challenges persist in reconciling representational expressiveness with real-time computational demands, developing formal safety verification methods for learned representations, and achieving robust generalization across diverse environmental conditions. Future research directions should prioritize hybrid modeling approaches that combine structured priors with neural networks, explainable representation learning techniques for critical scenario analysis, and standardized evaluation metrics that account for both performance and safety requirements. These advances will be pivotal in transitioning from research prototypes to deployable autonomous systems capable of operating reliably in complex real-world environments. Additionally, we list recent open-source work on implicit maps in Appendix.A.

\section{Future Trends}
Considering the aforementioned challenges and opportunities, we list some critical directions for future research as follows.

\subsection{Fully Automatic Lite Map}
End-to-end autonomous driving has not yet matured to the point where it can be commercially deployed at scale. As a result, explicitly designed lightweight maps remain the mainstream approach in the industry today. The number of related academic papers are also showing an increasing trend year by year, as summarized in Appendix.A. Looking ahead, a major research direction will continue to build upon the Lite map concept. Due to the fact that the automation rate in industry has always been low, the focus will be on maintaining high-quality map generation while achieving fully automated pipelines. The ultimate goal is to create a closed-loop system, where data collection, processing, and updates are seamlessly integrated without the need for any human intervention, enabling scalable and cost-effective map maintenance to support autonomous driving applications. 

\subsection{Zero-shot Map Learning}
Zero-shot learning (ZSL) is a machine learning approach where a model is able to recognize objects without having seen any labeled examples during training \citep{pourpanah2022review, chen2023zero}. Zero-shot learning is promising to complement online Lite map learning by enabling semantic flexibility, generalization to unseen environments, and reduced reliance on labeled data. Interesting research points might include:
\begin{enumerate}[label=\arabic*)]
\item
Zero-shot Map Element Recognition: Develop models that can identify rare or novel road features (e.g., new sign types, local lane conventions) purely through zero-shot transfer from pretrained vision–language models. 

\item
Reducing Label Dependency: Online Lite map learning often struggles with the need for high-quality, human-labeled data for training or validation. Zero-shot capabilities allow models to generalize to novel road structures or layouts without needing supervised fine-tuning.

\item
Uncertainty-aware ZSL for Mapping: Combine ZSL with uncertainty estimation to help determine when a newly observed feature is outside the model’s generalization range, improving safety and map reliability.
\end{enumerate}

\subsection{Implicit Map in VLA}
Vision-Language-Action (VLA) Model is an emerging paradigm in autonomous driving and embodied AI that aims to unify perception, reasoning, and control within a single model \citep{brohan2023rt, zhen20243d, kim2024openvla}. A VLA model takes visual inputs (e.g., camera images), language inputs (e.g., route commands, traffic rules, or instructions), and outputs corresponding driving actions (e.g., steering, throttle, braking). This architecture is inspired by the way humans interpret their surroundings, understand instructions, and interact with the environment in a goal-driven manner \citep{zhou2025autovla, zhou2025opendrivevla, jiang2025diffvla}.

Another concept in VLA is ``memory knowledge base'' or ``experience layer''. The purpose of this concept is to leverage the map experience stored in the cloud to assist the vehicle in better understanding driving scenarios. For example, when the vehicle approaches an intersection, can it retrieve similar intersections from the knowledge base to help predict possible routes, even under obstructed conditions?

VLA models naturally support the idea of implicit mapping: instead of relying on structured maps, they learn to associate visual cues and linguistic context with action policies. For instance, a VLA model might implicitly understand where an intersection or crosswalk is, or the layout of a parking lot, based solely on its multimodal training data, without ever using a formalized map structure. This makes implicit maps a byproduct or internal representation within a VLA model, offering a flexible and potentially scalable alternative to explicit maps.

\subsection{Foundation Model}
Foundation models are large-scale, pre-trained models that serve as a general-purpose backbone for a wide range of downstream tasks. Trained on massive and diverse datasets using self-supervised or weakly supervised learning, these models, such as GPT \citep{achiam2023gpt, mao2023gpt}, CLIP \citep{radford2021learning, yu2022coca}, or ViT \citep{kirillov2023segment, yang2024depth}, exhibit strong generalization, transferability, and the ability to adapt to new tasks with minimal additional data or fine-tuning. Foundation models can naturally support the formation and use of Implicit maps in multiple ways:

\begin{enumerate}[label=\arabic*)]
\item
Representation Learning: Foundation models learn latent spatial structures and semantic context from large-scale visual (and possibly LiDAR or multimodal) data, forming internal maps without the need for manual annotations or explicit formats.

\item
Multimodal Integration: Foundation models can combine vision, language, and temporal context, enabling them to reason about traffic layouts, navigational goals, or obstacles based on implicit cues, much like how humans understand an environment from limited inputs.

\item
Memory and Attention: Transformer-based foundation models use attention mechanisms to dynamically retrieve spatially and semantically relevant information, which can serve as a form of implicit localization and mapping.
\end{enumerate}

\section{Conclusions}

With the gradual adoption of autonomous driving, traditional HD map production methods have become inadequate to meet the frequency demands of urban road updates. Significant changes have taken place in maps. This article categorizes this evolution into three stages: HD maps, Lite maps, and Implicit maps. We provide a detailed elaboration on the characteristics of these three stages, the map production workflows, the technical challenges encountered, and relevant academic research. 

Despite the fact that many automakers used ``mapless autonomous driving" as a marketing gimmick in recent years, time has proven that their so-called ``mapless'' approach actually refers to driving without HD maps. Maps, however, continue to exist in other forms within the autonomous driving system and have never truly been absent. Lite maps and Implicit maps stand as the clearest evidence of this.
 
In the future, the development trend of maps will lean toward greater lightweight design. Cloud-based map databases will assist vehicle-side real-time perception, ultimately paving the way for end-to-end autonomous driving.

%%-------------------------------------------------------------------------
\section*{Acknowledgements}
The authors would like to acknowledge the support of the National Natural Science Foundation of China (grant No.: 42301503, 42171415).

\bibliographystyle{elsarticle-harv} 
\bibliography{bib/reference}

\clearpage
\appendix
\section{Related Opensourced Work}
\label{append:opensource}

\begin{table}[thtb]
\begin{center}
\footnotesize
\caption{Part of opensourced work for implicit map}
\begin{tabular}{c|c|p{4.5cm}}
\toprule
Year & Methods & Opensource Code \\
\midrule
\multirow{1}{*}{2019} & DisentPhys  & \url{github.com/TsuTikgiau/DisentPhys4VidPredict/} \\
\midrule
\multirow{1}{*}{2021} & COLA  & \url{github.com/google-research/google-research/tree/master/cola} \\
\midrule
\multirow{4}{*}{2022} & BEVFormer  & \url{github.com/HFAiLab/BEVFormer}	\\
 & NerfAcc  & \url{github.com/nerfstudio-project/nerfacc} \\
 & NeRFusion & \url{github.com/jetd1/NeRFusion} \\
 & SinNeRF & \url{github.com/VITA-Group/SinNeRF} \\
\midrule
\multirow{7}{*}{2023} & ExpQD & \url{github.com/codezakh/decomposition-0shot-vqa}	\\
 & SparseBEV  & \url{github.com/MCG-NJU/SparseBEV} \\
 & DynaVol  & \url{github.com/zyp123494/DynaVol} \\
 & LaRa & \url{github.com/valeoai/LaRa} \\
 & PanNeRF360  & \url{github.com/fuxiao0719/PanopticNeRF/tree/panopticnerf360} \\
 & NeRFDet & \url{github.com/facebookresearch/NeRF-Det} \\
 & OccFormer & \url{github.com/zhangyp15/OccFormer} \\
\midrule
\multirow{13}{*}{2024} & PotentialMP  & \url{github.com/devinluo27/potential-motion-plan-release} \\
 & OccWorld & \url{github.com/wzzheng/OccWorld} \\
 & IC-Mapper & \url{github.com/Brickzhuantou/IC-Mapper} \\
 & Neural MP  & \url{github.com/mihdalal/neuralmotionplanner} \\
 & Vit-Lens  & \url{github.com/TencentARC/ViT-Lens} \\
 & MBNSF  & \url{github.com/kavisha725/MBNSF} \\
 & BiGR  & \url{github.com/haoosz/BiGR} \\
 & LightningNeRF  & \url{github.com/VISION-SJTU/Lightning-NeRF} \\
 & NeuRAD & \url{github.com/georghess/neurad-studio} \\
 & GraphAD  & \url{github.com/zhangyp15/GraphAD} \\
 & DriveDreamer & \url{github.com/JeffWang987/DriveDreamer} \\
 & OccSora & \url{github.com/wzzheng/OccSora} \\
 & CurbSG  & \url{github.com/robot-learning-freiburg/CURB-SG} \\
\midrule
\multirow{1}{*}{2025} & GaussianWorld  & \url{github.com/zuosc19/GaussianWorld} \\
\bottomrule
\multicolumn{3}{l}{\footnotesize{* Updated by June 2025.}} \\
\end{tabular}
\end{center}
\end{table}

\begin{table}[thtb]
\begin{center}
\footnotesize
\caption{Part of opensourced work for online vectorization}
\begin{tabular}{c|c|p{4.5cm}}
\toprule
Year & Methods & Opensource Code \\
\midrule
\multirow{1}{*}{2022} & HDMapNet & \url{github.com/Tsinghua-MARS-Lab/HDMapNet} \\
\midrule
\multirow{9}{*}{2023} & CenterLineDet  & \url{github.com/TonyXuQAQ/CenterLineDet} \\
& VectorMapNet \ & \url{github.com/Mrmoore98/VectorMapNet_code} \\
& MapTR  & \url{github.com/hustvl/MapTR}	\\
& NMP & \url{github.com/Tsinghua-MARS-Lab/neural_map_prior} \\
& PivotNet  & \url{github.com/wenjie710/PivotNet} \\
& BeMapNet & \url{github.com/er-muyue/BeMapNet} \\
& MapVR & \url{github.com/ZhangGongjie/MapVR} \\
& TopoNet  & \url{github.com/OpenDriveLab/TopoNet} \\
& ScalableMap  & \url{github.com/jingy1yu/ScalableMap} \\
\midrule
\multirow{18}{*}{2024} & MapTRV2 & \url{github.com/hustvl/MapTR}	\\
& SMERF  & \url{github.com/NVlabs/SMERF} \\
& GeMap  & \url{github.com/cnzzx/GeMap} \\
& SatforHDMap  & \url{github.com/xjtu-cs-gao/SatforHDMap} \\
& LaneGAP  & \url{github.com/hustvl/LaneGAP} \\
& Topo2D  & \url{github.com/homothetic/Topo2D} \\
& TopoLogic  & \url{github.com/Franpin/TopoLogic} \\
& LaneGraph2Seq & \url{github.com/fudan-zvg/RoadNet} \\
& ADMap  & \url{github.com/hht1996ok/ADMap} \\
& CGNet & \url{github.com/XiaoMi/CGNet} \\
& StreamMapNet  & \url{github.com/yuantianyuan01/StreamMapNet} \\
& P-MapNet & \url{github.com/jike5/P-MapNet} \\
& MapQR & \url{github.com/HXMap/MapQR} \\
& SQD-MapNet & \url{github.com/shuowang666/SQD-MapNet} \\
& MapTracker & \url{github.com/woodfrog/maptracker} \\
& MGMap & \url{github.com/xiaolul2/MGMap} \\
& GenMapping & \url{github.com/lynn-yu/GenMapping} \\
& HRMapNet  & \url{github.com/HXMap/HRMapNet} \\
\midrule
\multirow{10}{*}{2025} & Uni-PrevPredMap  & \url{github.com/pnnnnnnn/Uni-PrevPredMap} \\
& PrevPredMap  & \url{github.com/pnnnnnnn/PrevPredMap} \\
& FastMap  & \url{github.com/hht1996ok/FastMap} \\
& MapFM & \url{github.com/LIvanoff/MapFM} \\
& SDTagNet  & \url{github.com/immel-f/SDTagNet} \\
& SuperMapNet  & \url{github.com/zhouruqin/SuperMapNet} \\
& UIGenMap & \url{github.com/xiaolul2/UIGenMap} \\
& MapDR  & \url{github.com/MIV-XJTU/MapDR} \\
& InstaGraM  & \url{github.com/juyebshin/InstaGraM} \\
& Chameleon & \url{github.com/XR-Lee/neural-symbolic} \\

\bottomrule
\multicolumn{3}{l}{\footnotesize{* Updated by June 2025.}} \\
\end{tabular}
\end{center}
\end{table}

\end{document}